\documentclass[10pt,twocolumn,letterpaper]{article}

\usepackage{iccv}              

\definecolor{iccvblue}{rgb}{0.21,0.49,0.74}
\usepackage[pagebackref,breaklinks,colorlinks,allcolors=iccvblue]{hyperref}

\usepackage{xspace}
\usepackage{xcolor}
\usepackage[font=small,skip=2pt]{caption}
\usepackage[normalem]{ulem}

\newcommand{\note}[1]{\color{black}{#1}\color{black}\xspace}
\newcommand{\bonote}[1]{\color{black}{#1}\color{black}\xspace}
\definecolor{isucolor}{RGB}{0,153,0} 
\newcommand{\isu}[1]{\color{black}{#1}\color{black}\xspace}
\newcommand{\jjm}[1]{\color{black}{#1}\color{black}\xspace}

\newcommand{\parlabel}[1]{\noindent\textbf{#1}}

\AtEndPreamble{
    \usepackage[capitalize]{cleveref}
    \crefname{equation}{Eq.}{Eqs.}      
    \Crefname{equation}{Eq.}{Eqs.}
    \crefname{algorithm}{Alg.}{Algs.}   
    \Crefname{algorithm}{Alg.}{Algs.}
    \crefname{section}{Sec.}{Secs.}     
    \Crefname{section}{Sec.}{Secs.}
    \crefname{table}{Tab.}{Tabs.}       
    \Crefname{table}{Tab.}{Tabs.}
    \crefname{figure}{Fig.}{Figs.}       
    \crefname{figure}{Fig.}{Figs.}       
    \crefname{appendix}{App.}{Apps.}       
    \crefname{appendix}{App.}{Apps.}       
}

\usepackage{multirow}

\def\paperID{11654} 
\def\confName{ICCV}
\def\confYear{2025}

\title{
On-device Sora: Enabling Training-Free Diffusion-based\\Text-to-Video Generation for Mobile Devices}

\author{
  Bosung Kim\thanks{Co-first author.},
  Kyuhwan Lee\thanks{Co-first author.},
  Isu Jeong,
  Jungmin Cheon,
  Yeojin Lee,
  Seulki Lee\thanks{Corresponding author.},
  \and
  \textnormal{Ulsan National Institute of Science and Technology} 
}

\begin{document}
\maketitle
\begin{abstract}

We present On-device Sora, the first model training-free solution for diffusion-based on-device text-to-video generation that operates efficiently on smartphone-grade devices. To address the challenges of diffusion-based text-to-video generation on computation- and memory-limited mobile devices, the proposed On-device Sora applies three novel techniques to pre-trained video generative models. First, Linear Proportional Leap (LPL) reduces the excessive denoising steps required in video diffusion through an efficient leap-based approach. Second, Temporal Dimension Token Merging (TDTM) minimizes intensive token-processing computation in attention layers by merging consecutive tokens along the temporal dimension. Third, Concurrent Inference with Dynamic Loading (CI-DL) dynamically partitions large models into smaller blocks and loads them into memory for concurrent model inference, effectively addressing the challenges of limited device memory. We implement On-device Sora on the iPhone 15 Pro, and the experimental evaluations show that it is capable of generating high-quality videos on the device, comparable to those produced by high-end GPUs. These results show that On-device Sora enables efficient and high-quality video generation on resource-constrained mobile devices. We envision the proposed On-device Sora as a significant first step toward democratizing state-of-the-art generative technologies, enabling video generation on commodity mobile and embedded devices without resource-intensive re-training for model optimization (compression). The code implementation is available at a GitHub repository\footnote{\label{github}\url{https://github.com/eai-lab/On-device-Sora}}.
\end{abstract}
\section{Introduction}

Recent advancements in generative models~\cite{suzuki2022survey} have significantly expanded capabilities in data generation across various modalities, including text~\cite{hu2017toward}, image~\cite{shaham2019singan}, and video~\cite{unterthiner2018towards}. In particular, diffusion-based models for image tasks~\cite{ho2020denoising, song2020denoising, peebles2023scalable, rombach2022high, saharia2022palette, dhariwal2021diffusion, zhang2023adding, podell2023sdxl} have emerged as foundational tools for a wide range of applications, such as image generation~\cite{ho2022cascaded}, image editing~\cite{kawar2023imagic}, and personalized content creation~\cite{zhang2024survey}. Further extending these technologies, diffusion-based generative models are now driving remarkable progress in video generation tasks~\cite{singer2022make, menapace2024snap, ho2022imagen, ho2022video, blattmann2023stable, molad2023dreamix, harvey2022flexible, blattmann2023align, voleti2022mcvd, wu2023tune, khachatryan2023text2video}, including video synthesis~\cite{liu2021generative} and real-time video analysis~\cite{orriols2004generative}. 

With the rapid expansion of mobile and embedded devices, there is an increasing demand for executing generative applications directly on-device. In response, extensive research has focused on developing on-device image generation methods~\cite{li2024snapfusion, vasu2023mobileone, castells2024edgefusion, choi2023squeezing, chen2023speed, zhao2023mobilediffusion}, with recent advancements extending to on-device video generation~\cite{yahia2024mobile, wu2024snapgen}. However, generating lightweight and compressed on-device video generation models necessitates additional model optimization (training), such as distillation~\cite{10.1145/3681758.3698013, lin2024animatediff, singer2024video}, quantization~\cite{tian2024qvd, zhao2024vidit}, and pruning~\cite{wu2024individual}, to reduce model size and computational complexity. Furthermore, mitigating performance degradation in the compressed models requires a time-intensive and iterative model re-training. This process usually demands substantial computational and memory resources, \eg, SnapGen-V~\cite{wu2024snapgen} takes over 150K training iterations on 256 NVIDIA A100 80GB GPUs to generate mobile-deployable compressed video generation models.

We introduce On-Device Sora, the first training-free framework that enables diffusion-based text-to-video generation on mobile devices, which allows video generative models to run directly on-device, enabling the production of high-quality videos on smartphone-grade devices. While previous approaches~\cite{wu2024snapgen} require additional training and substantial GPU resources to optimize video generation models for mobile environments, the proposed On-Device Sora eliminates the need for training and directly enables efficient on-device execution. Using pre-trained video generative models, \eg, Open-Sora~\cite{opensora}, \jjm{Pyramidal} Flow~\cite{jin2024pyramidal}, On-device Sora significantly enhances their efficiency, enabling on-device video generation with limited resources. 

To achieve this, we address key challenges of enabling on-device text-to-video generation: 1) Linear Proportional Leap (LPL) reduces the iterative denoising steps in the diffusion process by leaping through a significant portion of steps using the Euler’s method~\cite{biswas2013discussion} along an estimated direct trajectory, 2) Temporal Dimension Token Merging (TDTM) lowers computational complexity of STDiT (Spatial-Temporal Diffusion Transformer)~\cite{opensora} by merging consecutive tokens~\cite{bolya2022token,bolya2023token} in the attention layers, and 3) Conference Inference and Dynamic Loading (CI-DL) enables video generation with limited memory capacity of mobile devices by integrating concurrent model inference with the dynamic loading of models into memory. With these three proposed methods, high-quality video generation becomes feasible on smartphone-grade devices with limited computing resources, overcoming the requirements for substantial computational power, such as high-end GPUs. To the best of our knowledge, On-device Sora proposed in this work is the first training-free solution that enables the efficient generation of video directly on the device. The advantages of On-device Sora, \ie, directly deploying well-trained video generative models onto resource-constrained devices without time-consuming model modification and/or re-training, are expected to become even more pronounced with the active development of compact DiT-based text-to-video generative models~\cite{peebles2023scalable}.

We implement On-device Sora on the iPhone 15 Pro~\cite{apple2023} using Open-Sora~\cite{opensora}. The full implementation is available as open-source code in an anonymous GitHub\footref{github}. The extensive experiments are conducted to evaluate the performance of On-device Sora using the state-of-the-art video benchmark tool, \ie, VBench~\cite{huang2024vbench}, compared with NVIDIA A6000 GPUs. We also evaluate the proposed methods on additional text-to-video generative models, including Pyramidal Flow~\cite{jin2024pyramidal}. The experimental results demonstrate that On-device Sora maintains comparable video quality while accelerating generation speed with the proposed methods. While the iPhone 15 Pro has a GPU of 143 times less computational power~\cite{apple2023} and 16 times smaller memory compared to the NVIDIA A6000 GPU, the evaluation results show that On-device Sora significantly improves the efficiency of video generation by effectively compensating for the limited computing resources of the device.

\section{Background and Challenges}

We first provide a background of diffusion-based text-to-video generation (\eg Open-Sora~\cite{opensora}), which is the backbone of the proposed On-device Sora, and key challenges in realizing on-device video generation for mobile devices.

\subsection{Background: Diffusion-based Text-to-Video Generation}
\label{sec:background}


In general, diffusion-based text-to-video generation models, such as Open-Sora~\cite{opensora}, generate videos from user prompts (texts) through the three stages: 1) prompt (text) embedding, 2) latent video generation, and 3) video decoding.

\parlabel{1) Prompt (Text) Embedding.}
The first stage of diffusion-based text-to-video generation is to map a user prompt, a textual description of the desired video, to an embedding vector, which is used as input for the subsequent video generation stage. For example, to produce prompt embeddings from user texts, Open-Sora employs T5 (Text-to-Text Transformer)~\cite{raffel2020exploring}, a language model specifically fine-tuned to support video generation tasks.

\parlabel{2) Latent Video Generation.}
The next stage is to generate the latent video representation conditioned on the prompt embedding obtained from language models, \eg, T5 (Text-to-Text Transformer)~\cite{raffel2020exploring}. To this end, for instance, Open-Sora employs STDiT (Spatial-Temporal Diffusion Transformer)~\cite{opensora}, a diffusion-based text-to-video model using the Markov chain~\cite{zhang2023text}. Since maintaining temporal consistency across video frames is essential in video, STDiT~\cite{opensora} applies the spatial-temporal attention mechanism~\cite{yan2019stat} to the patch representations. It allows effective learning of temporal features across video frames through the temporal attention, enhanced by incorporating rope embeddings~\cite{su2024roformer}. During the forward process of STDiT, the Gaussian noise $\boldsymbol{\epsilon}_{k}$ is iteratively added to the latent video representation $\mathbf{x}_{k}$ over $K$ steps, transforming the intact video representation $\mathbf{x}_{0}$ into the complete Gaussian noise $\mathbf{x}_{K}$ in the latent space with the forward distribution $q(\mathbf{x}_{k}|\mathbf{x}_{k-1})$ as:
\begin{equation}
\begin{split}
    \mathbf{x}_{k} &= \sqrt{1-\beta_{k}} \mathbf{x}_{k-1} + \sqrt{\beta_{k}} \boldsymbol{\epsilon}_{k}
    \\
    q(\mathbf{x}_t | \mathbf{x}_{k-1}) &= \mathcal{N} (\mathbf{x}_{k} ; \sqrt{1-\beta_{k}} \mathbf{x}_{k-1},\beta_{k} \mathbf{I})    
\end{split}
    \label{eq:forward}
\end{equation}

\noindent where $\beta_{k}$ is the parameter determining the extent of noise. 

To generate the latent representation $\mathbf{x}_{0}$, the noise $\boldsymbol{\epsilon}_{k}$ is repeatedly removed (denoised) from the complete Gaussian noise $\mathbf{x}_{K}$ through the reverse process using the estimated noise with the reverse distribution $p_{\boldsymbol{\theta}}(\mathbf{x}_{k-1}|\mathbf{x}_t)$ modeled by STDiT~\cite{opensora} with the parameter set $\boldsymbol{\theta}$, as follows:
\begin{equation}
\begin{split}
    \mathbf{x}_{k-1} &= \boldsymbol{\mu}_{\boldsymbol{\theta}} (\mathbf{x}_{k}, k) + \sqrt{\beta_{k}} \boldsymbol{\epsilon} ~\text{ where}~ \boldsymbol{\epsilon} \sim \mathcal{N}(\mathbf{0}, \mathbf{I})
    \\
    p_{\boldsymbol{\theta}}(\mathbf{x}_{k-1}|\mathbf{x}_{k}) &= \mathcal{N}(\mathbf{x}_{k-1}; \boldsymbol{\mu}_{\boldsymbol{\theta}} (\mathbf{x}_{k}, k), \boldsymbol{\Sigma}_{\boldsymbol{\theta}} (\mathbf{x}_{k}, k))
\end{split}
    \label{eq:reverse}
\end{equation}

\noindent where $\boldsymbol{\mu}_{\boldsymbol{\theta}} (\mathbf{x}_{k}, k)$ and $\boldsymbol{\Sigma}_{\boldsymbol{\theta}} (\mathbf{x}_{k}, k)$ is the mean and variance of $\mathbf{x}_{k}$ estimated by STDiT training, respectively. This reverse process repeatedly denoises $\mathbf{x}_{k}$ into $\mathbf{x}_{k-1}$ over a large number of $1 {\leq} k {\leq} K$ denoising steps (\ie, from dozens to thousands of steps), eventually generating the de-noised latent video representation $\mathbf{x}_0$ close to the intact representation.

\begin{figure*}[!t]
    \centering
    \includegraphics[width=\linewidth]{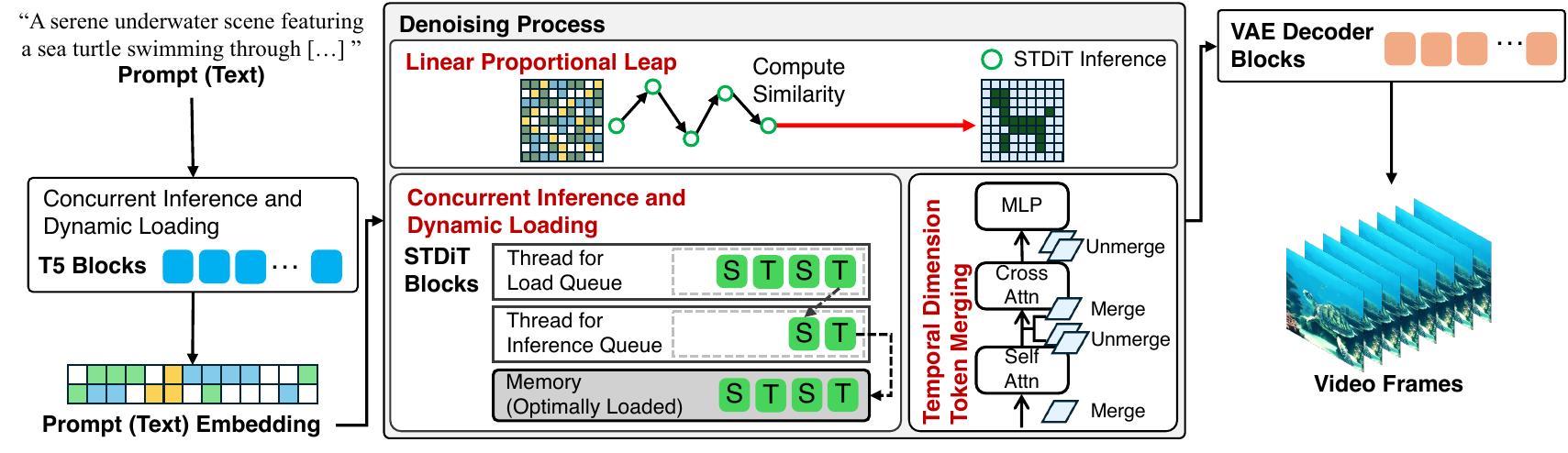}
    \caption{On-device Sora enables training-free text-to-video generation directly on the device by employing three key methods: 1) Linear Proportional Leap (LPL), 2) Temporal Dimension Token Merging (TDTM), and 3) Concurrent Inference with Dynamic Loading (CI-DL).}
    \label{fig:overview}
\end{figure*}

\parlabel{3) Video Decoding.}
Finally, the latent video representation $\mathbf{x}_0$ generated from STDiT is decoded and up-scaled into the human-recognizable video through the VAE (Variational Autoencoder)~\cite{doersch2016tutorial}. For example, the VAE employed in Open-Sora~\cite{opensora} utilizes both 2D and 3D structures; the 2D VAE is based on SDXL~\cite{podell2023sdxl}, while the 3D VAE adopts the architecture of MAGVIT-v2~\cite{yu2023language}. 

\subsection{Key Challenges in On-device Video Generation}
\label{sec:challenges}



\parlabel{C1) Excessive Denoising Steps.}
\Cref{tab:time-profiling} shows the latency of each model component in Open-Sora~\cite{opensora}, where the latent video generation process (denoising process) performed by STDiT is the most time-consuming. 
That is because a substantial number of denoising steps  
is required to remove the noise $\boldsymbol{\epsilon}_{k}$ from $\mathbf{x}_{K}$ to obtain $\mathbf{x}_{0}$ during latent video generation~\cite{song2020denoising}. Such extensive denoising iterations 
presents considerable challenges on mobile devices with constrained computational capabilities. 
While the numerous denoising steps are manageable in server-level environments—where the complete denoising process typically finishes within one minute—on mobile devices, it may take several tens of minutes. 
Accordingly, an effective approach to reduce denoising steps without model modification or re-training 
is essential to enable on-device video generation.

\begin{table}[!htb]
\renewcommand{\arraystretch}{1}
\caption{The number of executions (iterations) of each model component (\ie, T5~\cite{raffel2020exploring}, STDiT~\cite{opensora}, and VAE~\cite{doersch2016tutorial}) in Open-Sora~\cite{opensora} and their total latencies on the iPhone 15 Pro~\cite{apple2023}.
}
\label{tab:time-profiling}
\setlength\tabcolsep{5.5pt}
\resizebox{\columnwidth}{!}{%
\begin{tabular}{c|rrr}
\toprule[1pt] \hline
\textbf{Component} 
    & \multicolumn{1}{c}{\textbf{\begin{tabular}[c]{@{}c@{}} Iterations\end{tabular}}} 
    & \multicolumn{1}{c}{\textbf{\begin{tabular}[c]{@{}c@{}}Inference Time (s)\end{tabular}}} 
    & \multicolumn{1}{c}{\textbf{\begin{tabular}[c]{@{}c@{}}Total Latency (s) \end{tabular}}} \\ \hline
T5~\cite{raffel2020exploring}        &       1     &          110.505     &            110.505\\
STDiT~\cite{opensora}     &       50    &           35.366    &           1768.320\\
VAE~\cite{doersch2016tutorial}       &       1     &           135.047   &            135.047\\ \toprule[1pt]
\end{tabular}%
}
\end{table}


\parlabel{C2) Intensive Token Processing.}
While a large number of denoising steps poses a significant challenge to video generation on mobile devices, even a single denoising step itself is computationally intensive. The primary reason is that the computational complexity of the attention mechanism~\cite{niu2021review} in STDiT~\cite{opensora} grows quadratically with the token size, which significantly increases the computational load for token processing and, consequently, the model's inference latency. To address this challenge, Token merging~\cite{bolya2023token} has been proposed to improve the throughput of vision Transformer models, which progressively merges similar visual tokens within the transformer to accelerate the model inference latency by reducing the size of tokens to be processed. 
While token merging has been applied to diffusion models, it is only applied to spatial tokens~\cite{bolya2022token,bolya2023token} and has not been applied to the temporal tokens in video diffusion models, such as STDiT~\cite{opensora}. Thus, a novel token merging method is required to improve the computational efficiency of token processing within well-trained video generation models, while preserving high video quality.

\begin{figure}[!htb]
    \centering
    \includegraphics[width=\columnwidth]{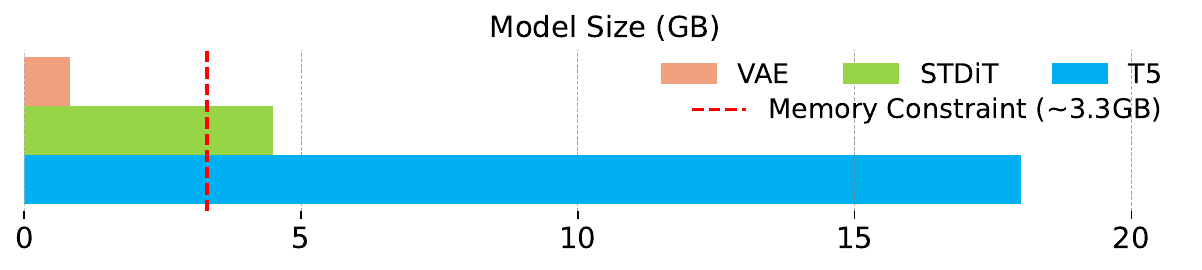}
    \caption{The size of Open-Sora models: T5~\cite{raffel2020exploring} (18.00 GB), STDiT~\cite{opensora} (4.50 GB), and {VAE~\cite{doersch2016tutorial} (0.82 GB)}, which exceeds the available memory capacity of the iPhone 15 Pro~\cite{apple2023} (3.3 GB).}
    \label{fig:peak-memory}
\end{figure}

\parlabel{C3) High Memory Requirements.}
\Cref{fig:peak-memory} shows the memory requirements of model components in Open-Sora~\cite{opensora}, \ie, VAE~\cite{doersch2016tutorial}, T5~\cite{raffel2020exploring}, and STDiT~\cite{opensora}, where their cumulative memory demand, \ie, 23 GB, surpasses the memory capacity of many mobile devices. For instance, the iPhone 15 Pro~\cite{apple2023}, with 8 GB of memory, restricts the available memory for a single application to 3.3 GB for system stability. Furthermore, the individual memory requirements of T5 and STDiT exceed 3.3 GB, creating challenges in loading them into memory. In addition, some memory must be reserved for model execution (inference), exacerbating memory shortages on mobile devices. Thus, limited device memory is another challenge that should be addressed to enable on-device video generation.

\section{Overview: On-device Sora}

\Cref{fig:overview} shows an overview of On-Device Sora, which takes Open-Sora~\cite{opensora} outlined in Sec. \ref{sec:background} as its backbone. On-device Sora enables efficient diffusion-based text-to-video generation on mobile devices by addressing three key challenges (Sec. \ref{sec:challenges}) through three proposed methods: 1) Linear Proportional Leap (LPL) (Sec. \ref{sec:ours1}), 2) Temporal Dimension Token Merging (TDTM) (Sec. \ref{sec:ours2}), and 3) Concurrent Inference with Dynamic Loading (CI-DL) (\Cref{sec:ours3}). These methods can also be applied to other diffusion-based text-to-video generation models, \eg, \jjm{Pyramidal} Flow~\cite{jin2024pyramidal}.

\section{Linear Proportional Leap}
\label{sec:ours1}

On-device Sora reduces the excessive number of denoising steps performed by STDiT~\cite{opensora} by introducing Linear Proportional Leap (LPL), which leverages the trajectory properties of Rectified Flow~\cite{liu2022flow}. It allows early stop of denoising steps through proportionally scaled linear leaps, without extra model training or modification of STDiT architecture.

\subsection{Rectified Flow}

The reverse diffusion process is performed through multiple denoising steps, transforming an initial Gaussian distribution into a desired distribution corresponding to the input prompt. Several ODE-based methods~\cite{zheng2023dpm, lu2022dpm} reformulate this process by training models to predict the drift at each time point, building the distribution trajectories from the initial to the target point by using ODE solvers~\cite{runge1895numerische, yang2023diffusion}.

Rectified Flow~\cite{liu2022flow} simplifies the transition from the initial point to the target point by training the model to predict a drift aligned with the direct linear trajectory connecting these two points. Using the Euler method~\cite{chen2018neural}, the $k$th trajectory $\boldsymbol{z_k}$ is derived by updating the previous trajectory $\boldsymbol{z_{k-1}}$ with the estimated drift $\boldsymbol{v}(P_k,t_k)$ and step size $dt_k$ defined by two sampled time steps $t_k$ and $t_{k+1}$, as follows:
\begin{equation}
\begin{split}
    &\boldsymbol{z}_k = \boldsymbol{z}_{k-1} + \boldsymbol{v}(P_k,t_k)dt_k ~~ \forall 1 \leq k \leq K
    \\
    & \text{where}~ t_{k} \in [0,1] ~\text{and}~ dt_k = 
\begin{cases} 
    t_k {-} t_{k+1}, & \text{if } t_k \neq t_K \\
t_k, & \text{if } t_k = t_K
\end{cases}
\end{split}
    \label{eq:rflow}
\end{equation}

\noindent Here, the time step $t_{k} {\in} [0,1]$ corresponds to the normalized reverse process at the $k$th denoising step, with $t_{k} {=} 1$ representing the time step at which data is fully noisy (start of denoising) and $t_{k} {=} 0$ corresponding to the time step when data reaches the desired distribution (end of denoising). The drift $\boldsymbol{v}(P_k,t_k)$ is predicted from STDiT~\cite{opensora} given the $k$th position on the trajectory, $P_{k} = t_k P_1 + (1 - t_k) P_0$, computed from linear interpolation with sampled time step, $t_{k}$~\cite{liu2022flow}.




In Rectified Flow~\cite{liu2022flow}, the model is trained to predict the direct direction toward target point at any point on the trajectory. This allows diffusion-based generation models for achieving a denoising process in few steps without significant performance degradation. 
With Rectified Flow~\cite{liu2022flow}, Open-Sora~\cite{opensora} generates video with $K=30$ or $50$ steps.

\subsection{Linear Proportional Leap}



\begin{figure}[!t]
    \centering
    \includegraphics[width=\linewidth]{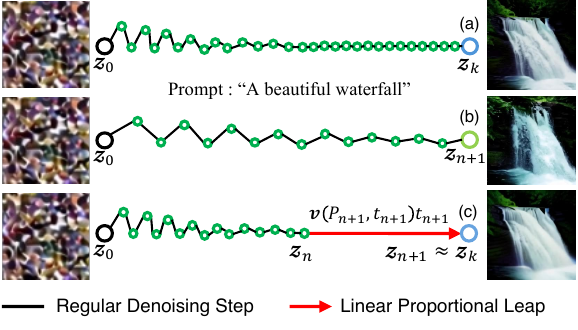}
    \caption{An abstracted illustration of trajectories and latent visualizations for $K=30$ and $n=15$: (a) Rectified Flow~\cite{liu2022flow} with full $k=30$ denoising steps, generating intact and complete video data, (b) Rectified Flow~\cite{liu2022flow} with $n+1=16$ denoising steps without applying Linear Proportional Leap, resulting in low-quality video data generation from variance with high step sizes ($dt_{k}$), and (c) Linear Proportional Leap with $n+1=15+1$ denoising steps, producing video data nearly equivalent to (a).}
    \label{fig:lpl-mainfig}
\end{figure}

Based on Rectified Flow~\cite{liu2022flow}, the proposed Linear Proportional Leap reduces denoising steps, as illustrated in \Cref{fig:lpl-mainfig}. 

If the $n$th data distribution is sufficiently close to the $K$th target distribution in the denoising process, the trajectories $\boldsymbol{z}_{n+1 ... K}$ would be approximately straight for the remaining time steps $t_{n+1 ... K}$, making the drift estimation $\boldsymbol{v}(P_k,t_k)dt_k$  in \Cref{eq:rflow} unnecessary for $k {>} n$. Consequently, $\boldsymbol{v}(P_k,t_k)dt_k$ is estimated only for $1 {\leq} k {\leq} n$, allowing the denoising process to stop early at the $(n+1)$th step, not to the full $K$th step. For the remaining steps $t_{n+1 ... K}$, the trajectories $\boldsymbol{z}_{n+1 ... K}$ linearly leap towards the target data distribution, with the straight direction of $\boldsymbol{v}(P_{n+1},t_{n+1})$ and $dt_{n+1}$ scaled proportionally to $t_{n+1 ... K}$.

By assuming $k=K$, $\boldsymbol{z}_{k}$ is derived from \Cref{eq:rflow} as:
\begin{equation}
\begin{split}
    \boldsymbol{z}_k &= \boldsymbol{z}_{k-1} + \boldsymbol{v}(P_k,t_k)dt_k
    \\
    &= \boldsymbol{z}_0 + \sum_{i=1}^{k-1} \boldsymbol{v}(P_i,t_i) ({t_i-t_{i+1}}) + \boldsymbol{v}(P_k, t_k) t_k
\end{split}
    \label{eq:rflow2}
\end{equation}

\noindent If the denoising process stops at the step $n$, the $n$th trajectory is represented as $\boldsymbol{z}_n = \boldsymbol{z}_0 + \sum_{i=1}^{n} \boldsymbol{v}(P_i,t_i) ({t_i-t_{i+1}})$. Then, if we apply the identical drift $\boldsymbol{v}(P_{n+1},t_{n+1})dt_{n+1}$ to the remaining $n+1 \leq i \leq k$ steps, \Cref{eq:rflow2} becomes:
\begin{equation}
\begin{split}
    \boldsymbol{z}_k &{=} \boldsymbol{z}_n {+} \boldsymbol{v}(P_{n+1},t_{n+1}) \sum_{i={n+1}}^{k-1}(t_i {-} t_{i+1}) {+} \boldsymbol{v}(P_{n+1},t_{n+1}) t_k
    \\
    &{=} \boldsymbol{z}_n {+} \boldsymbol{v}(P_{n+1},t_{n+1}) (t_{n+1} {-} t_{n+2} {+} \cdots {+} t_{k-1} {-} t_k {+} t_k)
    \\
    &{=} \boldsymbol{z}_n {+} \boldsymbol{v}(P_{n+1},t_{n+1}) t_{n+1}
\end{split}
    \label{eq:lpl}
\end{equation}

\noindent Thus, the required denoising steps are reduced to $n+1$ out of the total $k$ steps, allowing STDiT to be executed only $n+1$ times, with the last $(n+1)$th trajectory applied to its time step $t_{n+1}$, which enables Linear Proportional Leap.
Replacing $dt_{k}$ with $t_{n+1}$ in \Cref{eq:lpl}, instead of computing difference between the sampled time steps, can invoke an identical effect under assumption that later steps tend to sustain drift directions. 
It enables the immediate completion of the denoising process, as $t_{n+1}$ is equivalent to the remaining denoising steps required to reach the end of denoising.

\begin{figure} [!t]
  \centering
  \begin{minipage}[c]{0.51\columnwidth}
    \centering    
    \includegraphics[width=\columnwidth]{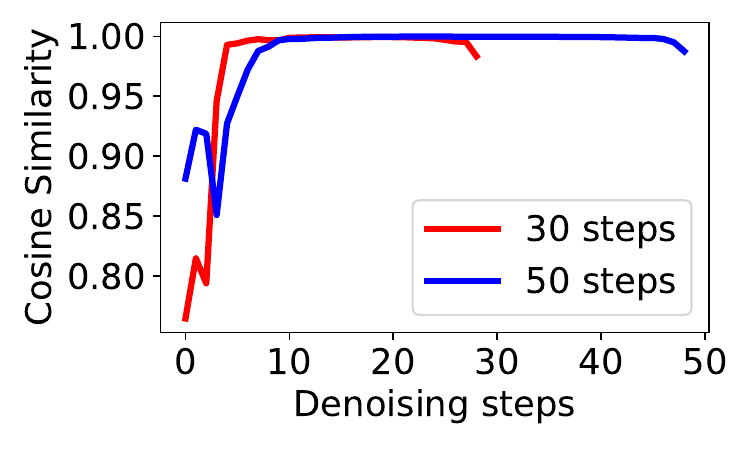}
  \end{minipage}%
  \begin{minipage}[c]{0.47\columnwidth}
  \caption{An example of cosine similarities between two adjacent drifts estimated from STDiT~\cite{opensora}, \ie, $\boldsymbol{v}(P_{n},t_{n})$ and $\boldsymbol{v}(P_{n-1},t_{n-1})$ for 30 (red) and 50 steps (blue).} 
    \label{fig:cos_siml}
  \end{minipage}
\end{figure}

Linear Proportional Leap can be dynamically applied by measuring the cosine similarity between two consecutive drifts $\boldsymbol{v}$ at runtime. When the cosine similarity appears that the current trajectory is sufficiently linear, a linear leap is made proportionally to the remaining steps. \Cref{fig:cos_siml} visualizes the cosine similarities between consecutive drifts, stabilizing after a certain number of steps, suggesting that the trajectory toward the target data distribution is nearly linear. 
This enables fewer steps by utilizing the leap with the larger step size to efficiently progress to the desired direction.

\section{Temporal Dimension Token Merging}
\label{sec:ours2}

On-device Sora reduces the computational complexity of the denoising process by introducing Temporal Dimension Token Merging (TDTM), which halves the size of tokens in STDiT~\cite{opensora} over the temporal dimension, decreasing the computation of self-attention quadratically and cross-attention in half. Unlike existing token merging applying to self-attentions over the spatial dimension, which exhibits suboptimal performance~\cite{bolya2022token,bolya2023token,feng2023efficient,li2024vidtome}, Temporal Dimension Token Merging leverages the temporal aspect of video frames to reduce computation while ensuring video quality.



\subsection{Token Merging}

STDiT~\cite{opensora} consists of multiple attention layers, \ie, cross- and self-attention, of the linear and quadratic complexity, respectively. In video generation, these attentions extend across two dimensions, \ie, spatial and temporal dimension. General model optimization techniques, \eg, pruning~\cite{reed1993pruning}, quantization~\cite{gray1998quantization}, and distillation~\cite{hinton2015distilling}, may reduce STDiT's attention computation. However, they necessitate model training (fine-tuning) or specialized hardware for implementation, and more importantly, the performance of video generation can hardly be preserved. In contrast, token merging~\cite{bolya2023token} reduces the token size processed in attentions, decreasing computational complexity without requiring model re-training or hardware-specific adaptations.

In STDiT~\cite{opensora}, the attention bias influenced by input tokens in cross-attention layers leads to higher computational demands than in self-attention layers.
As a result, developing an effective token merging method for cross-attention is crucial in video generation. However, existing token merging~\cite{bolya2022token, bolya2023token, li2024vidtome} applied to the cross-attention have shown suboptimal performance, and applying them to the self-attention in STDiT has observed video quality drops~\cite{bolya2022token,bolya2023token}.


\subsection{Temporal Dimension Token Merging}

\begin{figure}[!t]
    \centering
    \includegraphics[width=\linewidth]{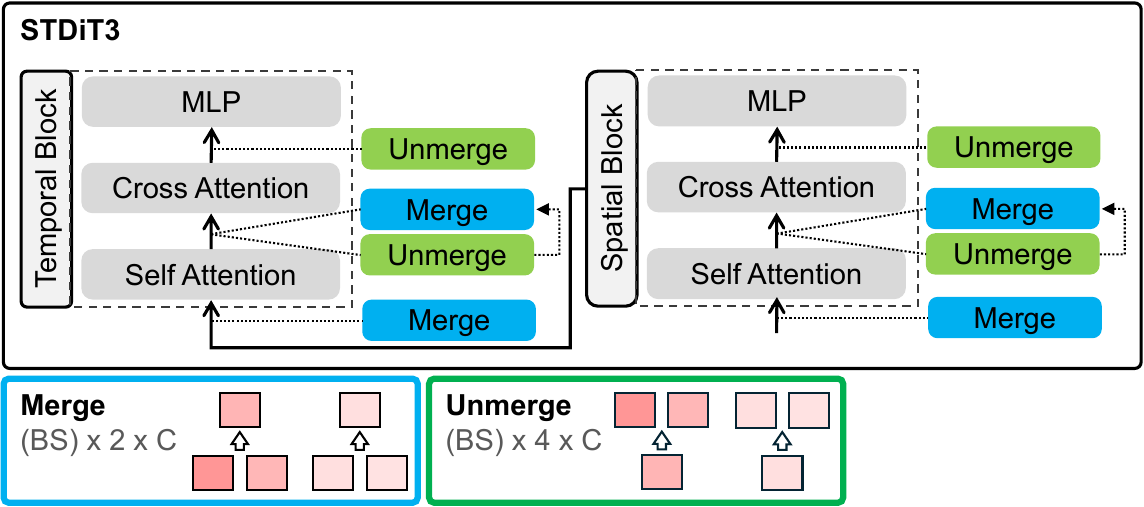}
    \caption{In attention layers of STDiT~\cite{opensora}, two consecutive tokens are merged along the temporal dimension and subsequently unmerged after processing, reducing the token size by half and the computational complexity up to a quarter.}
    \label{fig:TDTM-fig}
\end{figure}

\Cref{fig:TDTM-fig} illustrates Temporal Dimension Token Merging. Based on the hypothesis that successive video frames exhibit similar values, two consecutive frames are merged over the temporal dimension by averaging, creating a single token without the overhead of calculating frame similarity. This reduces the size of tokens by half while preserving the essential temporal information. Consequently, it decreases the computation of self-attention by a factor of four, according to the self-attention's quadratic complexity, $\mathcal{O}(n^{2})$. Similarly, it reduces the computation of cross-attention by half, based on its linear complexity, $\mathcal{O}(n m)$. Then, the output token processed through attentions replicates the dimensions for each frame, restoring them to their original size. 

Given the token $\boldsymbol{T}_{in}$ as an input for a self- or cross-attention with the dimension [$B$, $ST$, $C$], where $B$ denotes the batch size, $S$ is the number of pixel patches, $T$ is the number of frames, and $C$ is the channel dimension, the input token $\boldsymbol{T}_{in}$ is merged into $\boldsymbol{T}_{merged}$, with the index $i$, as:

\begin{gather}
    \boldsymbol{T}_{merged} = TDTM_{merge}(\boldsymbol{T}_{in})
    \\    
    TDTM_{merge}(\boldsymbol{T})[i] = \frac{1}{2} (\boldsymbol{T}[:,i,:] + \boldsymbol{T}[:,i+1,:])
\end{gather}

\noindent From this, two adjacent tokens are merged along the temporal dimension, producing the merged token $\boldsymbol{T}_{merged}$ of [$B$, $ST/2$, $C$], which is processed by self- or cross-attention.

After being processed by each attention, $\boldsymbol{T}_{merged}$ is unmerged into $\boldsymbol{T}_{unmerged}$ of the dimension [$B$, $ST$, $C$] as:

\begin{gather}
    \boldsymbol{T}_{unmerged} = TDTM_{unmerge}(\text{Attention}(\boldsymbol{T}_{merged}))
    \\
    TDTM_{unmerge}(\boldsymbol{T})[2i]= \boldsymbol{T}[:,i,:]
\end{gather}

\noindent where Attention$(\cdot)$ is either the self- or cross-attention.


\begin{figure*}[!t]
    \centering
    \includegraphics[width=2\columnwidth]{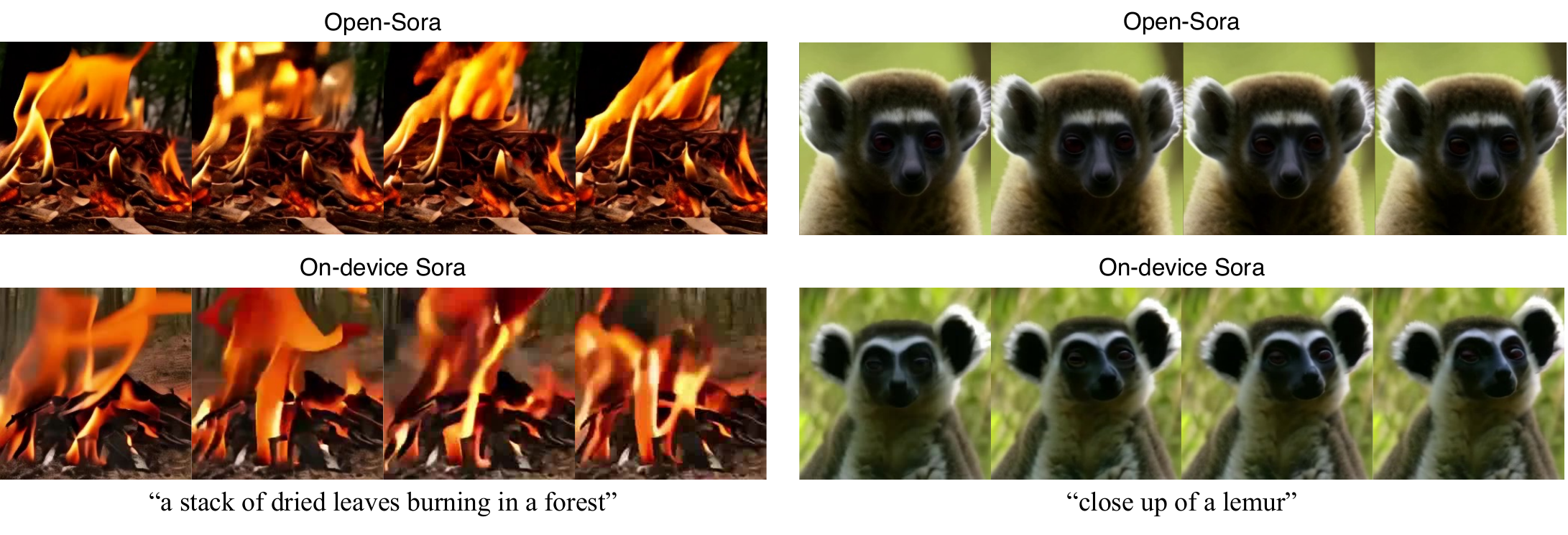}
    \caption{Example videos generated by On-device Sora and Open-Sora~\cite{opensora} (68 frames, 256×256).
    }
    \label{fig:end-to-end-generated-frames}
\end{figure*}

Temporal Dimension Token Merging can be selectively applied during the denoising process to minimize potential negative impacts on video quality that may arise from processing merged tokens. Specifically, out of a total of $K$ denoising steps, tokens can be merged only for the initial $k$ steps, while the tokens for the remaining $K{-}k$ steps remain unmerged. This is based on the observation that, when tokens are merged along the temporal dimension, the noise values vary slightly across frames—a phenomenon not observed in image diffusion~\cite{saharia2022palette} that does not involve a temporal dimension in the generation process. However, because noise values in the early denoising steps are less critical, it is expected that applying token merging exclusively for initial steps does not substantially drop video quality.

\section{Implementation}

We implement the proposed On-device Sora on iPhone 15 Pro~\cite{apple2023}, leveraging its GPU of 2.15 TFLOPS and 3.3 GB of available memory, with the two methods proposed in Sec. \ref{sec:ours1} and \ref{sec:ours2}. In addition, to execute large video generative models (\ie, T5 \cite{raffel2020exploring} and STDiT \cite{opensora}) with the limited device memory, we devise and implement Concurrent Inference with Dynamic Loading (CI-DL), which partitions the models into smaller blocks that can be loaded into the memory and executed concurrently. The details of CI-DL is described in \Cref{sec:ours3}. The model components—T5~\cite{raffel2020exploring}, STDiT~\cite{opensora}, and VAE~\cite{doersch2016tutorial}—in PyTorch~\cite{paszke2019pytorch} are converted to MLPackage, an Apple’s CoreML framework~\cite{sahin2021introduction} for machine learning apps. Since current version of CoreML \cite{apple2023} lacks support for certain diffusion-related operations in text-to-video generation, we develop custom solutions like xFormer \cite{xFormers2022} and cache-based acceleration. We implement denoising scheduling, sampling pipeline, and tensor-to-video conversion in Swift~\cite{swift} using Apple-provided libraries. To optimize models, T5~\cite{raffel2020exploring}, the largest in video generation, is quantized to int8, while others models  (STDiT~\cite{opensora} and VAE~\cite{doersch2016tutorial}) run in float32; we found that they are challenging to quantize due to sensitivity and performance degradation.

\section{Experiment}
\label{sec:experiment}
\isu{We evaluate the performance of On-device Sora on 68-frame videos at 256×256 resolution, using 800 text prompts sampled from VBench~\cite{huang2024vbench} and 1,000 each from Pandas70M~\cite{chen2024panda70m}, and VidGen~\cite{tan2024vidgen}. To assess both temporal and frame-level video quality, we utilize VBench~\cite{huang2024vbench}, the state-of-the-art benchmark for text-to-video generation, which provides comprehensive metrics, including subject consistency, background consistency, temporal flickering, motion smoothness, dynamic degree, aesthetic quality, and imaging quality.} Additional experiments on another text-to-video generation model, \ie, \jjm{Pyramidal} Flow~\cite{jin2024pyramidal}, are provided in 
\Cref{fig:lpl_pyramid} and ~\Cref{tab:LPL_pyramid} in  \Cref{sec:extra_experiment}.

\subsection{Video Generation Performance}
\label{sec:ex_video_generation_performance}

\begin{table*}[!t]
\setlength{\tabcolsep}{18pt}
\renewcommand{\arraystretch}{1.05}
\caption{The VBench~\cite{huang2024vbench} evaluation summary: On-device Sora vs. Open-Sora~\cite{opensora} (68 frames, 256×256), \note{full results in \Cref{tab:appendix-main-experiments} in \Cref{sec:extra_experiment}.}
}
\label{tab:main-experiments}
\resizebox{\textwidth}{!}{%
\begin{tabular}{l|ccccc|cc}
\toprule[1pt] \hline
  \multicolumn{1}{c|}{\multirow{2}{*}{\textbf{Method}}} &
  \multicolumn{5}{c|}{\textbf{Temporal Quality$\uparrow
$}} &
  \multicolumn{2}{c}{\textbf{Frame-Wise Quality$\uparrow
$}} \\ \cline{2-8} 
  \multicolumn{1}{c|}{} &
  \multicolumn{1}{c}{\begin{tabular}[c]{@{}c@{}}Subject \\ Consistency\end{tabular}} &
  \multicolumn{1}{c}{\begin{tabular}[c]{@{}c@{}}Background\\ Consistency\end{tabular}} &
  \multicolumn{1}{c}{\begin{tabular}[c]{@{}c@{}}Temporal\\ Flickering\end{tabular}} &
  \multicolumn{1}{c}{\begin{tabular}[c]{@{}c@{}}Motion\\ Smoothness\end{tabular}} &
  \multicolumn{1}{c|}{\begin{tabular}[c]{@{}c@{}}Dynamic\\ Degree\end{tabular}} &
  \multicolumn{1}{c}{\begin{tabular}[c]{@{}c@{}}Aesthetic\\ Quality\end{tabular}} &
  \multicolumn{1}{c}{\begin{tabular}[c]{@{}c@{}}Imaging\\ Quality\end{tabular}} \\ \hline

Open-Sora      & 	0.97& 	0.97& 	0.99& 	0.99& 	0.21& 	0.50& 	0.56    \\
On-device Sora & 	0.96& 	0.97& 	0.99& 	0.99& 	0.27& 	0.47& 	0.53    \\  \toprule[1pt]
                              
\end{tabular}%
}
\end{table*}



We evaluate the quality of videos generated by On-device Sora on the iPhone 15 Pro~\cite{apple2023}, in comparison to videos produced by Open-Sora~\cite{opensora} running on NVIDIA A6000 GPUs.
\note{\Cref{tab:main-experiments} summarizes the averaged results that compare those videos evaluated with VBench~\cite{huang2024vbench}, including categories such as animals, humans and lifestyle. \Cref{tab:appendix-main-experiments} and \Cref{fig:end-to-end-evaluation-2} in \Cref{sec:extra_experiment} provide detailed full per-category results.} The results demonstrate that On-device Sora generates videos with quality nearly equivalent to Open-Sora in most metrics, exhibiting only a slight drop in frame-level quality, averaging \isu{0.03}, while achieving an average \isu{0.06} improvement in dynamic degree.

\Cref{fig:end-to-end-generated-frames} shows example videos, compared with Open-Sora~\cite{opensora}. For the prompt ``a stack of dried leaves burning in a forest", both On-device Sora and Open-Sora generate visually plausible videos, illustrating stack of burning dry leaves and forest in the background. Similarly, for ``close-up of a lemur", both produce descriptive videos: On-device Sora shows a lemur turning its head, while Open-Sora delivers a less dynamic yet visually comparable depiction.

\begin{table*}[!t]
\setlength{\tabcolsep}{6.5pt}
\renewcommand{\arraystretch}{1.05}
\caption{The video quality and generation speedup under different settings of LPL (Linear Proportional Leap).}
\label{tab:early-stopping}
\resizebox{\textwidth}{!}{%
\small

\begin{tabular}{l|c|ccccc|cc|c}
\toprule[1pt] \hline
\multicolumn{1}{c|}{\multirow{2}{*}{\textbf{Dataset}}} &
  \multirow{2}{*}{\textbf{LPL Setting}} &
  \multicolumn{5}{c|}{\textbf{Temporal Quality$\uparrow $}} &
  \multicolumn{2}{c|}{\textbf{Frame-Wise Quality$\uparrow $}} &
  \multirow{2}{*}{\textbf{Speedup$\uparrow $}} \\ \cline{3-9}
\multicolumn{1}{c|}{} & 
   &
  \begin{tabular}[c]{@{}c@{}}Subject\\ Consistency\end{tabular} &
  \begin{tabular}[c]{@{}c@{}}Background\\ Consistency\end{tabular} &
  \begin{tabular}[c]{@{}c@{}}Temporal\\ Flickering\end{tabular} &
  \begin{tabular}[c]{@{}c@{}}Motion\\ Smoothness\end{tabular} &
  \begin{tabular}[c]{@{}c@{}}Dynamic\\ Degree\end{tabular} &
  \begin{tabular}[c]{@{}c@{}}Aesthetic\\ Quality\end{tabular} &
  \begin{tabular}[c]{@{}c@{}}Imaging\\ Quality\end{tabular} &
   \\ \hline
\multirow{3}{*}{VBench} 
                & Dynamic ($\mu$:17.73/30) & 0.97 & 0.97 & 0.99 & 0.99 & 0.20 & 0.50 & 0.56 & 1.53$\times$ \\ \cline{2-10} 
                & 16/30 (53\%)             & 0.97 & 0.97 & 0.99 & 0.99 & 0.18 & 0.50 & 0.55 & 1.94$\times$ \\
                & 30/30 (100\%)            & 0.97 & 0.97 & 0.99 & 0.99 & 0.21 & 0.50 & 0.57 & 1.00$\times$ \\ \toprule[1pt] 
\multirow{3}{*}{Pandas70M} 
                & Dynamic ($\mu$:17.66/30) & 0.97 & 0.97 & 0.99 & 0.99 & 0.22 & 0.47 &0.60  & 1.92$\times$ \\ \cline{2-10} 
                & 16/30 (53\%)             & 0.97 & 0.97 & 0.99 & 0.99 & 0.19  & 0.47 & 0.60 & 1.62$\times$ \\
                & 30/30 (100\%)            & 0.97 & 0.97 & 0.99 & 0.99 & 0.22 & 0.45 & 0.58 & 1.00$\times$ \\ \toprule[1pt] 
\multirow{3}{*}{VidGen} 
                & Dynamic ($\mu$:18.08/30) & 0.95 &  0.96& 0.98 & 0.99 & 0.36  & 0.47  & 0.57  & 1.94$\times$ \\ \cline{2-10} 
                & 16/30 (53\%)             & 0.95 & 0.96 & 0.98 & 0.98 & 0.33 & 0.46  & 0.56 & 1.53$\times$ \\
                & 30/30 (100\%)            & 0.96 & 0.97 & 0.99 & 0.99 & 0.39 & 0.48 & 0.57 & 1.00$\times$ \\ \toprule[1pt] 
\end{tabular}%
}
\label{tab:leap}
\end{table*}


\begin{table*}[!t]
\setlength{\tabcolsep}{8pt}
\renewcommand{\arraystretch}{1.05}
\caption{The video quality and speedup under different merging steps of TDTM (Temporal Dimension Token Merging). }
\label{tab:TDTM}
\resizebox{\textwidth}{!}{%
\small

\begin{tabular}{l|c|ccccc|cc|c}
\toprule[1pt] \hline
\multicolumn{1}{c|}{\multirow{2}{*}{\textbf{Dataset}}} &
  \multirow{2}{*}{\textbf{Merging Steps}} &
  \multicolumn{5}{c|}{\textbf{Temporal Quality$\uparrow $}} &
  \multicolumn{2}{c|}{\textbf{Frame-Wise Quality$\uparrow $}} &
  \multirow{2}{*}{\textbf{Speedup$\uparrow $}} \\ \cline{3-9}
\multicolumn{1}{c|}{} & 
   &
  \begin{tabular}[c]{@{}c@{}}Subject\\ Consistency\end{tabular} &
  \begin{tabular}[c]{@{}c@{}}Background\\ Consistency\end{tabular} &
  \begin{tabular}[c]{@{}c@{}}Temporal\\ Flickering\end{tabular} &
  \begin{tabular}[c]{@{}c@{}}Motion\\ Smoothness\end{tabular} &
  \begin{tabular}[c]{@{}c@{}}Dynamic\\ Degree\end{tabular} &
  \begin{tabular}[c]{@{}c@{}}Aesthetic\\ Quality\end{tabular} &
  \begin{tabular}[c]{@{}c@{}}Imaging\\ Quality\end{tabular} &
   \\ \hline

\multirow{3}{*}{VBench} 
                & 30/30 (100\%)            & 0.97 & 0.97 & 0.99 & 0.99 & 0.06 & 0.50 & 0.56 & 1.27$\times$ \\
                & 15/30 (50\%)            & 0.97 & 0.97 & 0.99 & 0.99 & 0.12 & 0.50 & 0.57 & 1.13$\times$ \\ 
                & 0/30 (0\%)               & 0.96 & 0.97 & 0.99 & 0.99 & 0.23 & 0.50 & 0.58 & 1.00$\times$ \\ \toprule[1pt] 
\multirow{3}{*}{Pandas70M} 
                & 30/30 (100\%)           & 0.98 & 0.97 & 0.99 & 0.99 & 0.09 & 0.48 & 0.59  & 1.23$\times$ \\
                & 15/30 (50\%)            & 0.98 & 0.97 &  0.99 & 0.99 & 0.13 & 0.47 & 0.58 & 1.11$\times$ \\ 
                & 0/30 (0\%)               & 0.97 & 0.97 & 0.99 & 0.99 & 0.22 & 0.45 & 0.58 & 1.00$\times$ \\ \toprule[1pt] 
\multirow{3}{*}{VidGen} 
                & 30/30 (100\%)            & 0.97 & 0.97 & 0.99 & 0.99 & 0.14 & 0.50 & 0.59 & 1.70$\times$ \\
                & 15/30 (50\%)            & 0.97 & 0.98 & 0.99 & 0.99 & 0.16 & 0.47 & 0.58  & 1.36$\times$ \\ 
                & 0/30 (0\%)               & 0.96 & 0.97 & 0.99 & 0.99 & 0.39 & 0.48 & 0.57 & 1.00$\times$ \\ \toprule[1pt] 
\end{tabular}%
}
\end{table*}

\subsection{Linear Proportional Leap}

\Cref{tab:early-stopping} presents the video generation performance and speedup of On-device Sora when applying Linear Proportional Leap (LPL) proposed in Sec. \ref{sec:ours1}. In the table, `LPL Setting' indicates the number of denoising steps used for video generation out of a total of 30 steps, while the remaining steps are omitted by LPL. We also evaluate a dynamic version of LPL, referred to as `Dynamic' in \Cref{tab:early-stopping}, which determines the number of denoising steps at runtime based on the cosine similarities between two adjacent drifts estimated using STDiT~\cite{opensora}. The dynamic LPL halts denoising when a predefined number of cosine similarity measurements fail to improve beyond a $10^{-4}$ tolerance, after a designated minimum number of denoising steps (50\%).

Overall, LPL enables video generation with quality comparable to Open-Sora~\cite{opensora}, without noticeable visual degradation (\eg, 0.743 vs. 0.736 in average of VBench~\cite{huang2024vbench}),
while accelerating video generation up to $1.94\times$ without any model optimization or re-training. 
LPL reduces denoising latency linearly without extra computation, enabling efficient video generation while retaining robust performance.

\begin{figure}[!htb]
    \centering
    \includegraphics[width=\columnwidth]{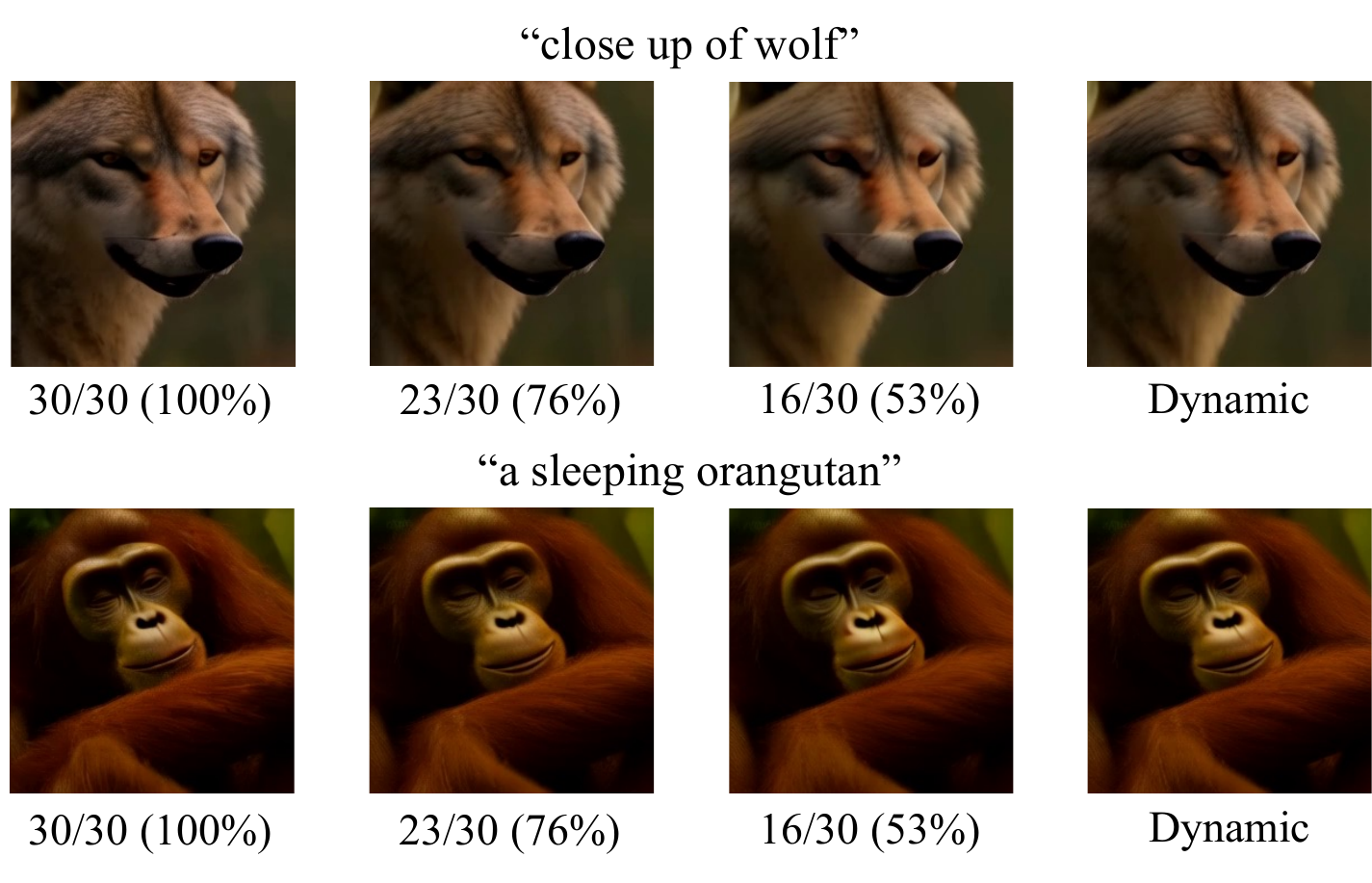}     
    \caption{The snapshots of videos (68 frames, 256×256 resolution) applied with various LPL settings (\Cref{tab:early-stopping}).}
    \label{fig:lpl-end-to-end-generated-frames1}
\end{figure}

\Cref{fig:lpl-end-to-end-generated-frames1} presents example videos generated using LPL, where all LPL settings consistently produce semantically identical target videos, with most video quality remaining stable across various numbers of denoising steps.

\subsection{Temporal Dimension Token Merging}

\Cref{tab:TDTM} presents the video quality and video generation speedup achieved when varying numbers of denoising steps to which Temporal Dimension Token Merging (TDTM) (Sec.~\ref{sec:ours2}) is applied, indicated as `Merging Steps' in the table, out of a total of 30 denoising steps. 
The results indicate that increasing the number of merging steps consistently accelerates video generation, ranging from $1.11\times$ to $1.70\times$ speedups, while maintaining stable quality metrics; the average scores for VBench remain at 0.736.
Nonetheless, some declines in the dynamic degree metric are observed, revealing a trade-off between maintaining visual dynamics and reducing token processing complexity. This indicates the importance of striking a balance between video dynamics and speedup. We found that selectively applying TDTM to specific denoising steps can effectively reduce visual noises. For instance, limiting TDTM to the first 15 denoising steps while not applying it to the rest of the denoising steps tends to result in a less severe quality drop compared to applying TDTM to all steps, which can mitigate issues like flickering or dynamic degree to provide improved video quality.

\begin{figure}[!htb]
    \centering
    \includegraphics[width=\columnwidth]{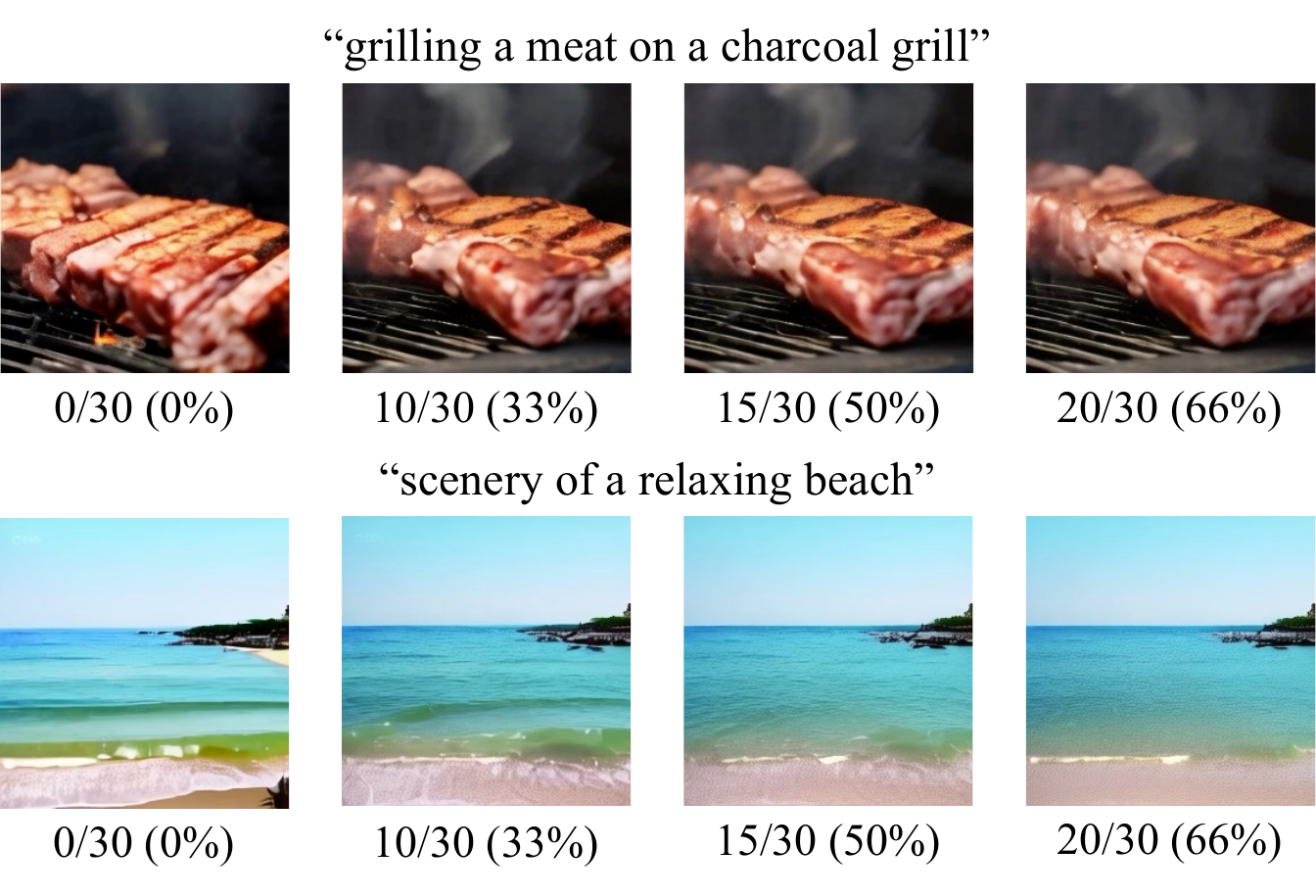}
    \caption{The snapshots of videos (68 frames, 256×256 resolution) applied with various merging steps of TDTM (\Cref{tab:TDTM}).} 
    \label{fig:tdtm-generated-frames}
\end{figure}

\Cref{fig:tdtm-generated-frames} shows examples of video frames generated with varying numbers of denoising steps to which TDTM is applied, out of 30 steps. The results show that even as TDTM is applied to an increasing number of denoising steps, the quality of the video frames seems to remain consistent.

\subsection{Video Generation Latency}

\Cref{tab:execution-time} shows video generation latencies of two resolutions when each of the proposed methodologies—LPL, TDTM, and CI-DL—is applied individually, as well as the latency when all of them are applied together (`All'). Latencies are measured with LPL activated at the 15th denoising step and TDTM applied throughout all steps, reported as the mean of three independent experiments. `STDiT' and `Total' specify whether the latency is measured solely for STDiT~\cite{opensora} or for end-to-end video generation, including T5~\cite{raffel2020exploring} and VAE~\cite{doersch2016tutorial}. The results demonstrate substantial latency reductions for each methodology compared to the case without using the proposed methods, \ie, 293.51 vs. 1768.32 seconds (\Cref{tab:time-profiling}) for STDiT. For the 192$\times$192 resolution video, STDiT (denoising process) takes less than five minutes when all three methodologies are applied. Additionally, it indicates that the methodologies do not interfere with each other, instead work synergistically to enhance latency.



\begin{table}[!htb]
\setlength{\tabcolsep}{6.5pt}
\renewcommand{\arraystretch}{1.1}
\caption{Ablation study on video generation latency (s). `All' denotes the combined application of LPL, TDTM, and CI-DL.}
\label{tab:execution-time}
\small
\resizebox{\columnwidth}{!}{%
\begin{tabular}{c|c|rrrrr}
\toprule[1pt] \hline
\textbf{Resolution} & \textbf{Measurement} & \multicolumn{1}{c}{\textbf{LPL}} & \multicolumn{1}{c}{\textbf{TDTM}} & \multicolumn{1}{c}{\textbf{CI-DL}} & \multicolumn{1}{c}{\textbf{All}} \\ \hline
\multirow{2}{*}{192$\times$192} & STDiT & 390.72 & 696.03 & 566.81 & 293.51 \\ 
                                & Total & 514.24 & 823.03 & 691.25 & 416.50 \\ \hline
\multirow{2}{*}{256$\times$256} & STDiT & 573.88 & 965.40 & 947.67 & 454.48 \\
                                & Total & 754.08 & 1148.63 & 1127.88 & 638.09 \\ \hline             
\toprule[1pt]
\end{tabular}
}
\end{table}

\section{Related Work}

\parlabel{On-device Video Generation.}
On-device video generation has recently gained attention due to the growing demand for real-time, efficient, and privacy-preserving content creation, and recently begun to be studied. MobileVD~\cite{yahia2024mobile}, based on UNet~\cite{ronneberger2015u} in Stable Video Diffusion~\cite{blattmann2023stable}, reduces computational costs by employing pruning, including low-resolution fine-tuning, temporal multi-scaling, and optimizations in channel and time blocks. SnapGen-V~\cite{wu2024snapgen} utilizes Stable Diffusion v1.5~\cite{rombach2022high} for processing image information while efficiently handling temporal information through Conv3D and attention-based layers. Furthermore, with fine-tuning, the denoising step is reduced. However, these approaches require substantial GPU resources for model optimization, \eg, MobileVD gone 100K training iterations on four A100 GPUs, and SnapGen-V utilized more than 150K iterations on 256 NVIDIA A100 GPUs. In contrast, On-Device Sora instantly applies without additional model training or optimization, removing the requirement for GPU resources.

\parlabel{Rectified Flow.}
While Open-Sora~\cite{opensora} reduces the number of denoising steps by leveraging Rectified Flow~\cite{liu2022flow}, most related approaches~\cite{esser2024scaling,zhu2025slimflow} require conditioned model training and/or distillation~\cite{zhu2025slimflow}. In contrast, the proposed Linear Proportional Leap effectively reduces the denoising steps without a significant performance drop, as validated using VBench~\cite{huang2024vbench}, without requiring model re-training or distillation. Notably, it can be easily activated at runtime by just calculating the cosine similarities of drifts between consecutive denoising steps. 

\parlabel{Token Merging.}
Most token merging methods~\cite{bolya2022token,bolya2023token,feng2023efficient} primarily focus on image generation, where tokens are merged based on the spatial similarity rather than temporal similarity. Although some temporal token merging have been proposed, they are applied to models in other domains~\cite{gotz2024efficient,li2024vidtome, chen2024tempme}, not in video generation. As such, Temporal Dimension Token Merging is the first to apply token merging based on the successive similarities between frames in video generation. Additionally, while previous works apply token merging to self-attention due to performance degradation~\cite{bolya2022token,bolya2023token,feng2023efficient,li2024vidtome}, On-device Sora shows that token merging can be applied to cross-attention with minimal performance loss, achieving 50\% merging ratio.

\section{Conclusion}

We propose On-device Sora, the first training-free solution for generating videos on mobile devices using diffusion-based models. It addresses key challenges in video generation to enable efficient on-device operation. The issue of extensive denoising steps is addressed through Linear Professional Leap, the challenge of handling large tokens is mitigated with Temporal Dimension Token Merging, and memory limitations are overcome through Concurrent Inference with Dynamic Loading. These proposed efficient on-device video generation methodologies are not limited to On-device Sora but are broadly applicable to various applications, providing advancements in on-device video generation without additional model re-training.
{
    \small
    \bibliographystyle{ieeenat_fullname}
    \bibliography{bibliography}
}

\clearpage
\appendix

\section{Appendix - Implementation}
\subsection{Concurrent Inference with Dynamic Loading}
\label{sec:ours3}


On-device Sora tackles the challenge of limited device memory in text-to-video generation, which restricts the on-device inference of large diffusion-based models, by introducing Concurrent Inference with Dynamic Loading (CI-DL), which partitions models and executes them in a concurrent and dynamic manner.

\parlabel{Concurrent Inference}
The model components of On-device Sora~\cite{liu2024sora}, \ie, STDiT~\cite{opensora} and T5~\cite{raffel2020exploring}, easily exceed the available memory of many mobile devices, \eg, 3.3 GB RAM of iPhone 15 Pro, as shown in \Cref{fig:peak-memory}. Given that the Transformer architecture~\cite{wolf2020transformers}, which is the backbone for both T5~\cite{raffel2020exploring} and STDiT~\cite{opensora}, we partition these models into smaller blocks (segments) and load them into memory accordingly for model inference.

\begin{figure}[!ht]
    \centering
    \includegraphics[width=\columnwidth]{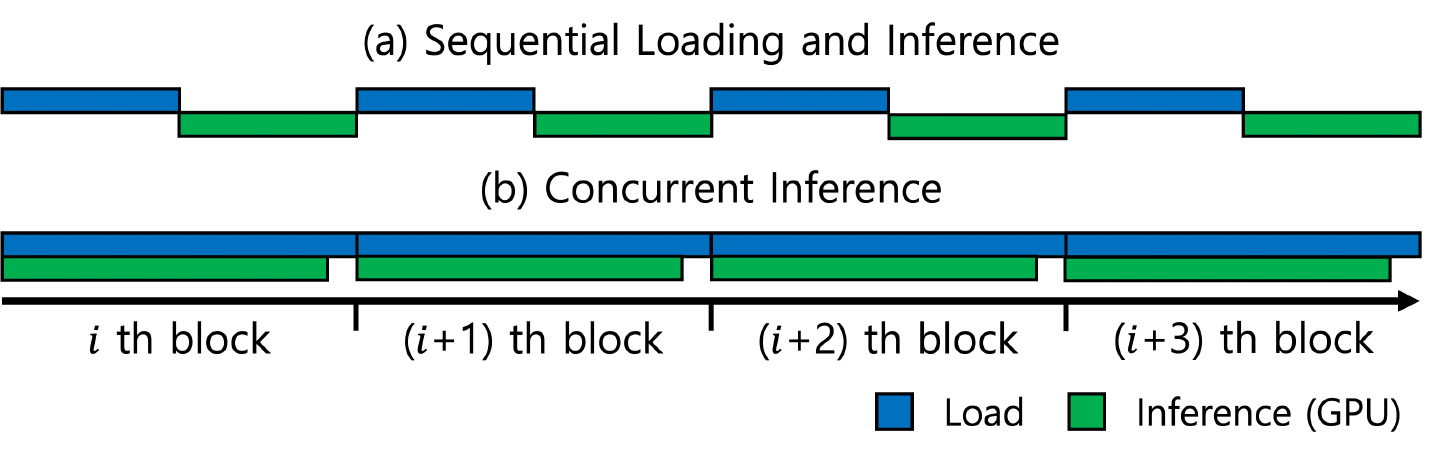}
    \caption{The block loading and inference cycles for (a) sequential loading and inference, and (b) concurrent inference. 
    }
    \label{fig:Partitioned_CI_STDiT}
\end{figure}

To execute video model inference using the model partitioning, each block must be loaded onto memory sequentially before execution, increasing the overall latency of video generation by incurring block loading time. \Cref{fig:Partitioned_CI_STDiT}-(a) shows the sequential block load and inference cycle of STDiT~\cite{opensora}, where GPU remains idle intermittently, waiting for each block to be loaded into memory, and only begins execution after the loading is complete. This sequential loading and execution process significantly increases the overall latency of model inference.

To minimize the increase in model inference latency caused by sequential block loading and execution, we propose Concurrent Inference, which leverages both the CPU and GPU for parallel block loading and execution; CPU loads the $(i+1)$th block, while GPU concurrently executes the $i$th block. Initially, the first and second blocks are loaded into memory concurrently, with the first block completing its loading first. Subsequently, the inference of the first block and the loading of the second block occur simultaneously. This process continues such that the inference of the $i$th block and the loading of the $(i+1)$th block overlap, ensuring continuous parallelism until the final block. \Cref{fig:Partitioned_CI_STDiT}-(b) depicts the load and inference cycle of STDiT with Concurrent Inference, which shows that GPU is active without almost no idle time by performing block loading and inference in parallel.

Given the number of model blocks $b$, block loading latency $l$, and inference latency of block $e$, the latency reduction $r$ achieved through Concurrent Inference is given by:
\begin{equation}
    r = b \cdot \min (l, e) - \alpha
    \label{eq:concurrent_inference}
\end{equation}
where $\alpha$ is the overhead caused by the block loading. 

Given the large number of denoising steps performed by STDiT, which is partitioned into multiple blocks for execution, similar to T5~\cite{raffel2020exploring}, the number of blocks $b$ is expected to be large, leading to a significant reduction in latency. It is expected to accelerate the overall model inference effectively regardless of the device's memory capacity. When the available memory is limited, then $b$ increases, while with larger memory, both $l$ and $e$ increase. In either case, it can result in an almost constant latency reduction $r$ in \Cref{eq:concurrent_inference}.


\parlabel{Dynamic Loading}
To further enhance the model inference latency, we propose Dynamic Loading, which is applied in conjunction with Concurrent Inference. It maintains a subset of model blocks in memory without unloading them, with the selection for the subset of blocks to be retained in memory dynamically determined based on the device's available memory at runtime.

The available memory of the device can vary at runtime based on the system status and configurations of applications running on the mobile device. By retaining a subset of model blocks in memory, the overhead of reloading these blocks during subsequent steps of Concurrent Inference can be eliminated, enabling reductions in model inference latency. To achieve this, we measure the device's run-time memory capacity and the memory required for inferring a single model block during the initial step of Concurrent Inference. Next, the memory allocated for retaining certain model blocks is dynamically determined as the difference between the available memory and the memory required for inferring a model block. Then, a series of model blocks that fit within this allocated memory is loaded in a retained state.

\begin{figure}[!ht]
    \centering
    \includegraphics[width=\columnwidth]{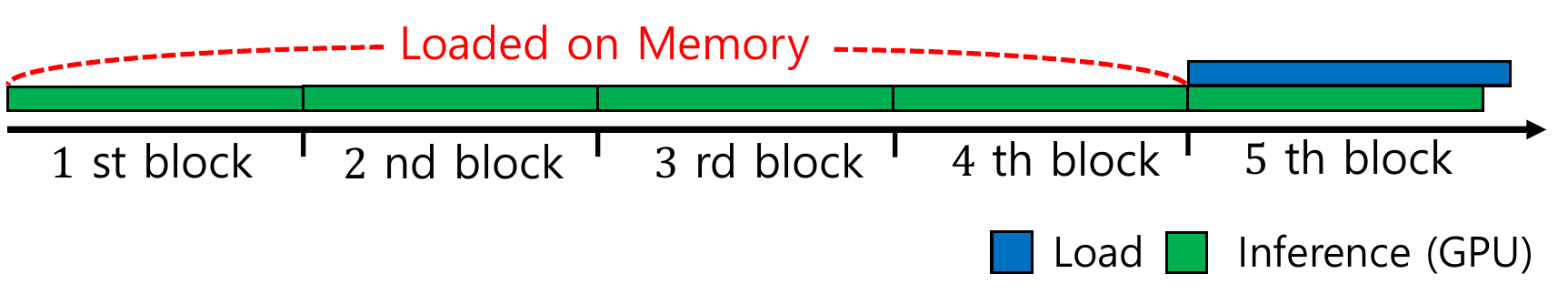}
    \caption{The block loading and inference cycle for Dynamic Loading applied with Concurrent Inference.}
    \label{fig:partitioning}
\end{figure}

\Cref{fig:partitioning} depicts Dynamic Loading; the first four model blocks are loaded in a retrained state. Unlike other blocks, \eg, the 5th block, these blocks are not unloaded to memory after the initial step, reducing block loading overhead. 

\begin{table*}[!htb]
\setlength{\tabcolsep}{12pt}
\renewcommand{\arraystretch}{1.1}
\caption{The VBench~\cite{huang2024vbench} evaluation by category: On-device Sora vs. Open-Sora~\cite{opensora} (68 frames, 256×256 resolution).
}
\label{tab:appendix-main-experiments}
\resizebox{\textwidth}{!}{%
\begin{tabular}{l|l|ccccc|cc}
\toprule[1pt] \hline
\multicolumn{1}{c|}{\multirow{2}{*}{\textbf{Category}}} &
  \multicolumn{1}{c|}{\multirow{2}{*}{\textbf{Method}}} &
  \multicolumn{5}{c|}{\textbf{Temporal Quality$\uparrow
$}} &
  \multicolumn{2}{c}{\textbf{Frame-Wise Quality$\uparrow
$}} \\ \cline{3-9} 
\multicolumn{1}{c|}{} &
  \multicolumn{1}{c|}{} &
  \multicolumn{1}{c}{\begin{tabular}[c]{@{}c@{}}Subject \\ Consistency\end{tabular}} &
  \multicolumn{1}{c}{\begin{tabular}[c]{@{}c@{}}Background\\ Consistency\end{tabular}} &
  \multicolumn{1}{c}{\begin{tabular}[c]{@{}c@{}}Temporal\\ Flickering\end{tabular}} &
  \multicolumn{1}{c}{\begin{tabular}[c]{@{}c@{}}Motion\\ Smoothness\end{tabular}} &
  \multicolumn{1}{c|}{\begin{tabular}[c]{@{}c@{}}Dynamic\\ Degree\end{tabular}} &
  \multicolumn{1}{c}{\begin{tabular}[c]{@{}c@{}}Aesthetic\\ Quality\end{tabular}} &
  \multicolumn{1}{c}{\begin{tabular}[c]{@{}c@{}}Imaging\\ Quality\end{tabular}} \\ \hline
\multirow{2}{*}{Animal}       & Open-Sora      & 	0.97& 	0.98& 	0.99& 	0.99& 	0.15& 	0.51& 	0.56 \\
                              & On-device Sora & 	0.95& 	0.97& 	0.99& 	0.99& 	0.28& 	0.48& 	0.55    \\ \hline
\multirow{2}{*}{Architecture} & Open-Sora      & 	0.99& 	0.98& 	0.99& 	0.99& 	0.05& 	0.53& 	0.60\\
                              & On-device Sora & 	0.98& 	0.98& 	0.99& 	0.99& 	0.12& 	0.49& 	0.56    \\\hline
\multirow{2}{*}{Food}         & Open-Sora      & 	0.97& 	0.97& 	0.99& 	0.99& 	0.26& 	0.52& 	0.60\\
                              & On-device Sora & 	0.95& 	0.97& 	0.99& 	0.99& 	0.38& 	0.48& 	0.53    \\ \hline
\multirow{2}{*}{Human}        & Open-Sora      & 	0.96& 	0.97& 	0.99& 	0.99& 	0.38& 	0.48& 	0.57\\
                              & On-device Sora & 	0.96& 	0.96& 	0.99& 	0.99& 	0.43& 	0.48& 	0.55    \\ \hline
\multirow{2}{*}{Lifestyle}    & Open-Sora      & 	0.97& 	0.97& 	0.99& 	0.99& 	0.23& 	0.45& 	0.56\\
                              & On-device Sora & 	0.96& 	0.97& 	0.99& 	0.99& 	0.25& 	0.45& 	0.53    \\ \hline
\multirow{2}{*}{Plant}        & Open-Sora      & 	0.98& 	0.98& 	0.99& 	0.99& 	0.15& 	0.50& 	0.58\\
                              & On-device Sora & 	0.97& 	0.98& 	0.99& 	0.99& 	0.16& 	0.46& 	0.55        \\ \hline
\multirow{2}{*}{Scenery}      & Open-Sora      & 	0.98& 	0.98& 	0.99& 	0.99& 	0.10& 	0.50& 	0.50\\
                              & On-device Sora & 	0.97& 	0.98& 	0.99& 	0.99& 	0.17& 	0.48& 	0.47        \\ \hline
\multirow{2}{*}{Vehicles}     & Open-Sora      & 	0.97& 	0.97& 	0.99& 	0.99& 	0.37& 	0.48& 	0.54\\
                              & On-device Sora & 	0.94& 	0.96& 	0.98& 	0.99& 	0.44& 	0.47& 	0.49 \\  \toprule[1pt]
\end{tabular}%
}
\end{table*}

\begin{figure*}[!t]
    \centering
    \includegraphics[width=\linewidth]{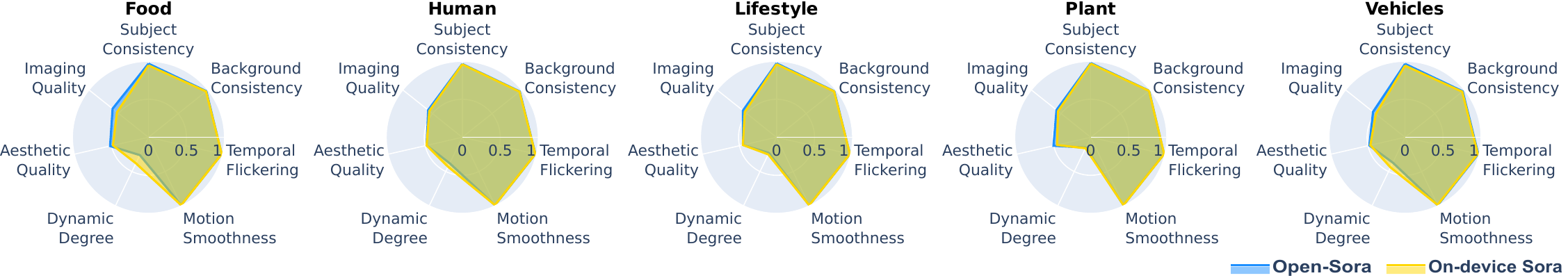}
    \caption{A visual comparison of videos generated by On-device Sora and Open-Sora~\cite{opensora}, evaluated using VBench~\cite{huang2024vbench}.}
    \label{fig:end-to-end-evaluation-2}
\end{figure*}

Applying Dynamic Loading, the latency reduction $r$ in \Cref{eq:concurrent_inference} for Concurrent Inference is updated as:
\begin{equation}
    r = b \cdot \min (l, e) + d \cdot \max (0, l - e)  - \alpha \cdot (1 - d / b)
    \label{eq:dynamic loading-2}
\end{equation}
where $d$ is the number of blocks maintained in memory. As the number of model blocks retained in memory increases (\ie, a larger $d$), the overhead of loading \and unloading blocks is further minimized, fully accelerating the overall model inference under the device's run-time memory constraints. Dynamically keeping some model blocks in memory with a retrained state is particularly advantageous for STDiT~\cite{opensora} that is iteratively executed for denoising steps, which entails loading and reusing the same blocks for each step.

\section{Appendix - Additional Experiments}
\label{sec:extra_experiment}

\subsection{Video Generation Performance}

{In \Cref{sec:ex_video_generation_performance}, we evaluate the quality of videos generated on an iPhone 15 Pro~\cite{apple2023} and compare them to videos produced by Open-Sora running on NVIDIA A6000 GPUs using VBench~\cite{huang2024vbench} as the evaluation benchmark. \isu{The evaluation is conducted on 68-frame videos at 256×256 resolution, using text prompts provided by VBench~\cite{huang2024vbench}, consisting of 100 examples each across eight categories: animals, architecture, food, humans, lifestyle, plants, scenery, and vehicles. \Cref{tab:appendix-main-experiments} and \Cref{fig:end-to-end-evaluation-2} present the comprehensive comparison of generated videos on all categories.}

\subsection{Concurrent Inference with Dynamic Loading}

\label{sec:CIDL}

\Cref{fig:exp_partitioned_stdit}-(b) illustrates the model block loading and inference cycles of STDiT~\cite{opensora} when applying the proposed Concurrent Inference (Sec. \ref{sec:ours3}), whereas \Cref{fig:exp_partitioned_stdit}-(a) depicts the case without its application. It can be observed that, with Concurrent Inference, the GPU executes each model block for inference without almost no idle time in parallel with the model block loading, indicated by the overlap between the red (loading) and black (inference) boxes in \Cref{fig:exp_partitioned_stdit}-(b), resulting in a block inference latency reduction from 29s to 23s. In contrast, without Concurrent Inference, each model block inference is executed only after the corresponding block is fully loaded to memory, indicated by the lack of overlap between the red (loading) and black (inference) boxes in \Cref{fig:exp_partitioned_stdit}-(a). Consequently, the total latency of the full denoising process using multiple executions of STDiT~\cite{opensora} is reduced by approximately 25\%, decreasing from 1,000 to 750 seconds for 30 denoising steps. Given that STDiT is executed multiple times to perform numerous denoising steps, it significantly accelerates the STDiT's overall inference. In addition, when applied with Dynamic Loading (Sec. \ref{sec:ours3}), it achieves an additional average speed improvement of 17 seconds as reloading is not required for certain model blocks that are retained in memory, provided in \Cref{eq:dynamic loading-2}.

\begin{figure} [!htb]
    \centering
    \includegraphics[width=\linewidth]{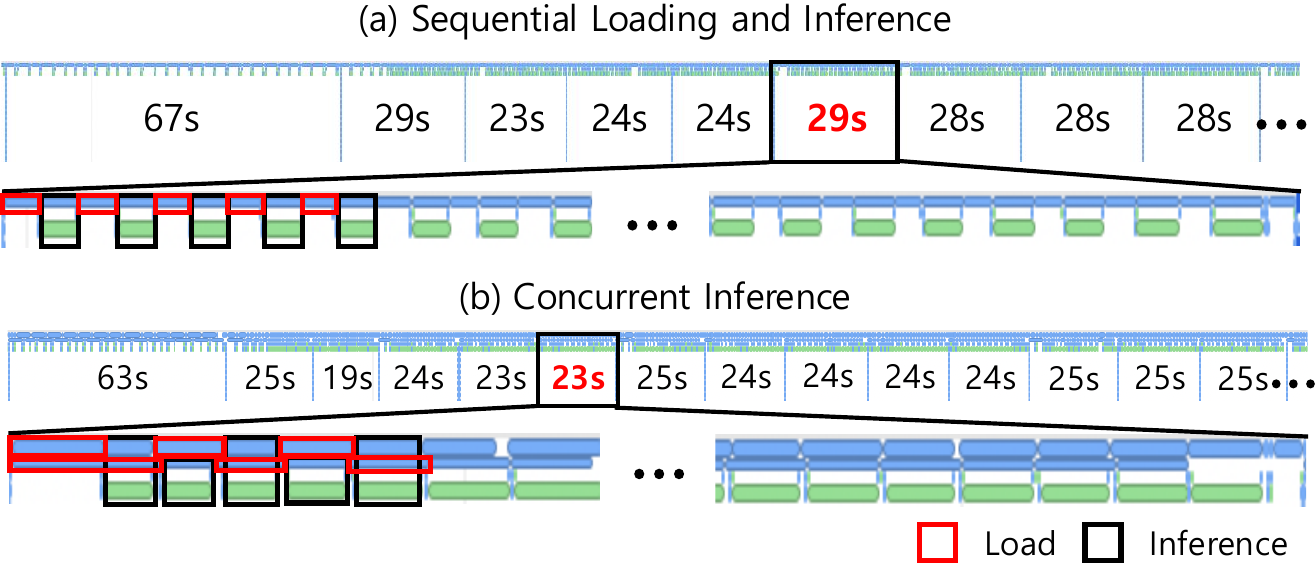}
    \caption{The block loading and inference cycles of STDiT \cite{opensora} without (a) and with (b) Concurrent Inference. The red box represents the loading cycle, while the black box indicates the model block inference on the iPhone 15 Pro's GPU.
    }
    \label{fig:exp_partitioned_stdit}
\end{figure}

\begin{figure} [!ht]
    \centering
    \includegraphics[width=\linewidth]{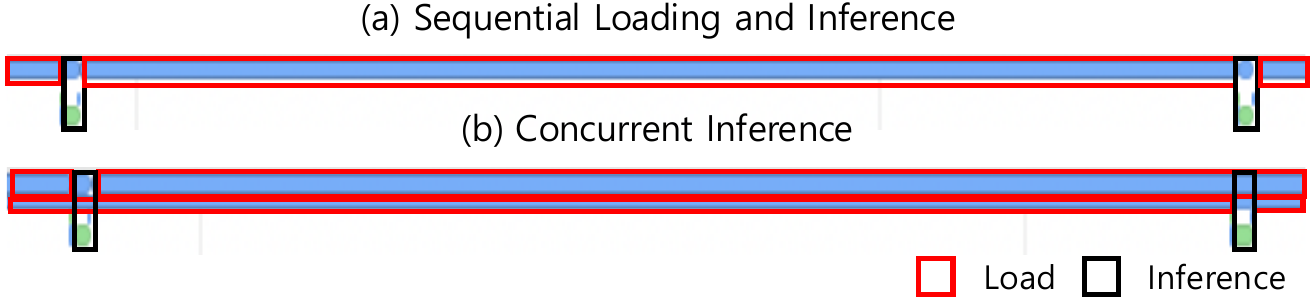}
    
    \caption{The block loading and inference cycles of T5~\cite{raffel2020exploring} without (a) and with (b) Concurrent Inference. 
    }
    \label{fig:exp_partitioned_ci_t5}
\vspace{-4pt}
\end{figure}

\Cref{fig:exp_partitioned_ci_t5} shows the case of T5~\cite{raffel2020exploring}. Unlike STDiT, which exhibits similar latencies for both block loading and inference execution, T5 exhibits a much longer block loading latency relative to inference latency. Consequently, the overlap between the model loading and inference is minimal, as shown by the small region of overlap between the red (loading) and black (inference) boxes. As a result, the latency improvement is expected to be less substantial, as in \Cref{eq:concurrent_inference}. Nevertheless, the inference latency is reduced from 164 to an average of 137 seconds, achieving a reduction of 16\%. This result implies the effectiveness of concurrent loading and inference, even in cases when there is an imbalance between the latencies of model block loading and inference.

\subsection{Ablation Study of Linear Proportional Leap (LPL) on Other Text-to-Video Models}
\bonote{The primary constraint for applying Linear Proportional Leap (LPL) is that the model must be a flow-matching–based model trained with a rectified flow target~\cite{liu2022flow}. Since} \jjm{Pyramidal} \bonote{Flow~\cite{jin2024pyramidal}, one of the state-of-the-art open-source video generation models, satisfies this condition, we conduct experiments on it.} 


\bonote{In a manner similar to Open-Sora with Rectified Flow~\cite{liu2022flow},} \jjm{Pyramidal} \bonote{Flow consistently demonstrates a high degree of straightness in its generation process following the initial generation unit (as illustrated in \Cref{fig:cos_sim_pyramid}). As demonstrated in \Cref{tab:LPL_pyramid}, the application of LPL during generation reduces the overall generation time by nearly 38\%, decreasing from 264.60 seconds to 165.15 seconds. Notably,} \jjm{Pyramidal} \bonote{Flow employs a cascading model procedure in which subtle variances are treated as noise to be subsequently denoised. Since these variances become negligible in the final output, Linear Proportional Leap is particularly well-suited to this architecture. Our experiments involving multiple prompts (as shown in \Cref{fig:lpl_pyramid}) confirmed that removing nearly half of the total denoising steps does not compromise performance while preserving both visual quality and temporal consistency. Furthermore, we observed that additional steps at later units can also be omitted due to the high straightness of} \jjm{Pyramidal} \bonote{Flow. Future work will further investigate this phenomenon by exploring the compatibility of flow-based models with the current LPL algorithm, developing the algorithm to be more compatible to other type of models.}

\begin{table}[!htb]
\setlength{\tabcolsep}{3pt}
\resizebox{\columnwidth}{!}{
\begin{tabular}{c|r|c|c|c}
\hline\hline
\textbf{Models} & \textbf{Time (s)} & \textbf{Resolution} & \textbf{DiT Steps} & \textbf{Speedup} \\ \hline\hline
\jjm{Pyramidal} Flow - 2B & 262.03 & 1280 $\times$ 768 & 270 & 1.00$\times$ \\ \hline
\jjm{Pyramidal} Flow - 2B - LPL (2) & 165.15 & 1280 $\times$ 768 & 159 & 1.59$\times$  \\ \hline
\jjm{Pyramidal} Flow - 2B & 48.88& 640 $\times$ 384 & 270 & 1.00$\times$ \\ \hline
\jjm{Pyramidal} Flow - 2B - LPL (2) & 30.54& 640 $\times$ 384  & 159 &1.60$\times$  \\ \hline
\hline
\end{tabular}
}
\caption{Ablation study conducted on \jjm{Pyramidal} Flow~\cite{jin2024pyramidal} based on the miniFLUX~\cite{flux2024} architecture. `DiT steps' represent the total number of DiT forward computations required for the denoising process. Speedups are measured relative to the corresponding model, with execution time computed as the average of three independent runs. Notably, LPL does not degrade video quality, even when using nearly half of its total steps.}
\label{tab:LPL_pyramid}
\end{table}

\begin{figure}[!htb]
    \centering
    \includegraphics[width=\linewidth]{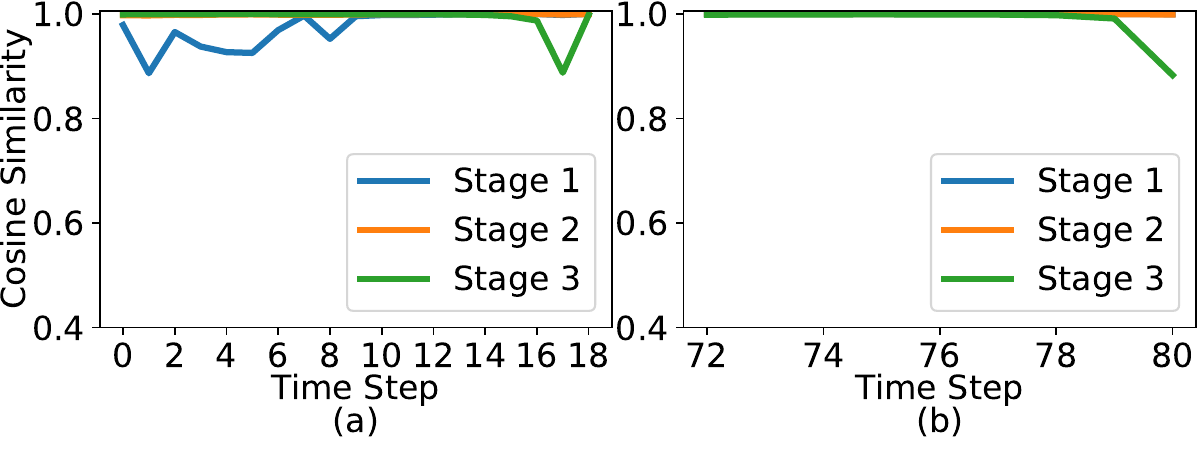}
    \caption{Cosine similarities between adjacent drifts estimated by \jjm{Pyramidal} Flow~\cite{jin2024pyramidal} at each stage. (a) and (b) show cosine similarities computed at the first and last units, respectively. The values approach 1.0 at the last unit across all stages. Computations follow the experimental settings in~\cite{jin2024pyramidal}, using 20 denoising steps in (a) and 10 in (b).}
    \vspace{-8pt}
    \label{fig:cos_sim_pyramid}    
\end{figure}

\begin{figure*}
    \centering
    \includegraphics[width=\linewidth]{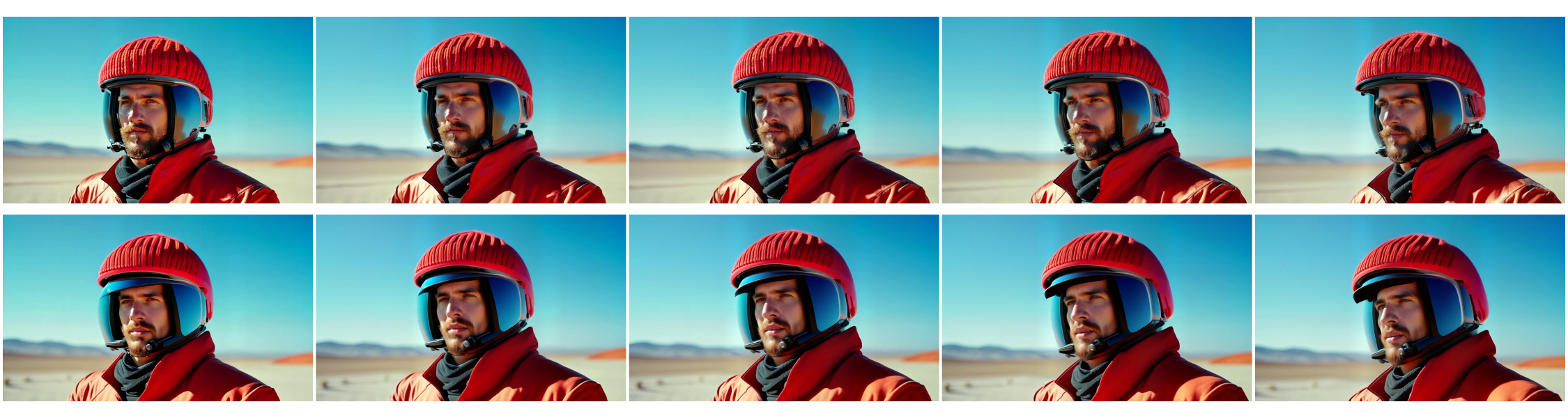}
    \caption*{Prompt: \textit{"A movie trailer featuring the adventures of a 30-year-old spaceman wearing a red wool-knitted motorcycle helmet, blue sky, salt desert, cinematic style, shot on 35mm film, vivid colors."}}
    \includegraphics[width=\linewidth]{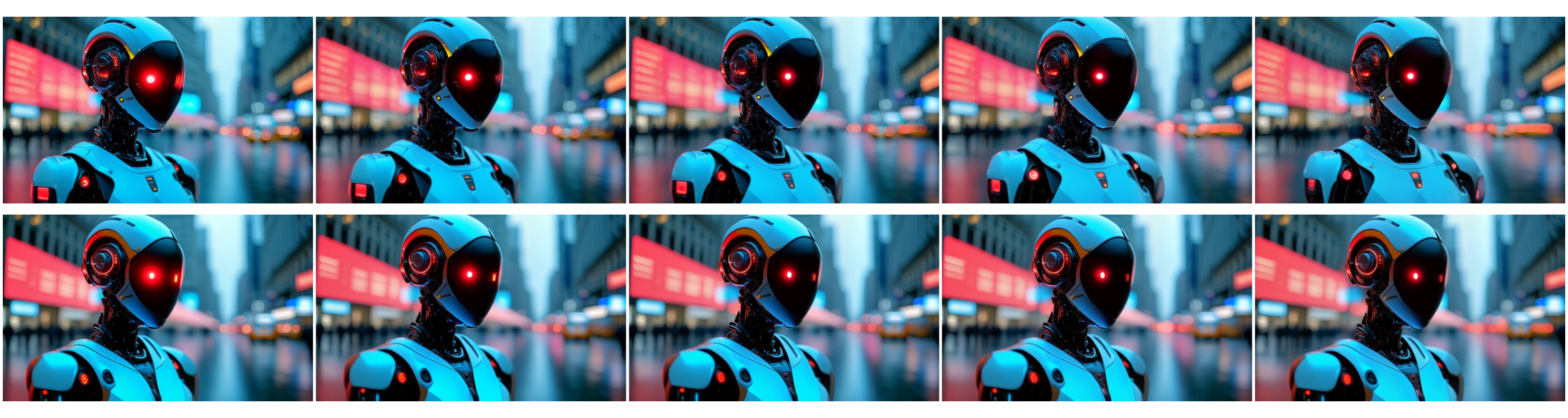}
    \caption*{Prompt: \textit{"A cybernetic humanoid scans the streets with red eyes as holographic screens flicker around its head, displaying futuristic data."}}
    \includegraphics[width=\linewidth]{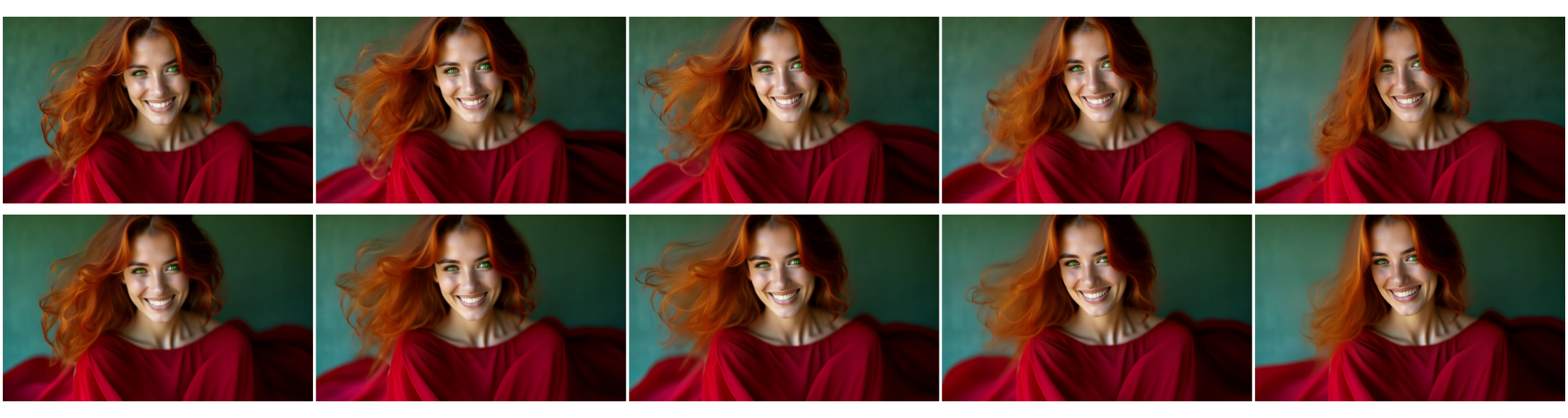}
    \caption*{Prompt: \textit{"A green-eyed woman with auburn hair and a mysterious smile, wearing a flowing red dress."}}
\caption{Videos generated using naive \jjm{Pyramidal} Flow~\cite{jin2024pyramidal} (top) and \jjm{Pyramidal} Flow with the proposed Linear Proportional Leap (bottom) under identical prompts. Linear Proportional Leap (LPL) reduces the total number of denoising steps by nearly half \note{(\ie, from 270 to 159)}, while maintaining video generation results that are highly comparable to those achieved using the full set of denoising steps and preserving most details.}
    \label{fig:lpl_pyramid}
\vspace{-8pt}
\end{figure*}

\section{Appendix - Discussions and Limitations}
\label{sec:app_dis_lim}

\parlabel{Latency Improvement.}
Although On-device Sora enables efficient video generation, the generation latency remains higher compared to utilizing high-end GPUs; it requires several minutes to generate a video, whereas an NVIDIA A6000 GPU takes one minute. This discrepancy is evident due to the substantial disparity in computational resources between them. For instance, the iPhone 15 Pro's GPU features up to 2.15 TFLOPS with 3.3 GB of available memory, compared to the NVIDIA A6000, which offers up to 309 TFLOPS and 48 GB of memory. Despite this significant resource gap, On-device Sora achieves exceptional efficiency in video generation. Currently, it utilizes only the iPhone 15 Pro's GPU. We anticipate that the latency could be significantly enhanced if it can leverage NPU (Neural Processing Unit), \eg, the iPhone 15 Pro's Neural Engine~\cite{apple2023}, which delivers a peak performance of 35 TOPS. However, the current limitations in Apple's software and SDK support for state-of-the-art diffusion-based models~\cite{opensora} make the iPhone's NPU challenging to utilize effectively. 
We look forward to the development of software support for NPUs and leave the exploration of this for future work. Also, we plan to investigate the potential of NPUs on a variety of mobile devices, such as Android smartphones.

\parlabel{Model Optimization. 
}
In On-device Sora, only T5~\cite{raffel2020exploring} is quantized to int8, reducing its size from 18 GB to 4.64 GB, while STDiT~\cite{opensora} and VAE~\cite{doersch2016tutorial} are executed with float32 due to their performance susceptibility, which has a significant impact on video quality. Additionally, we do not apply pruning~\cite{reed1993pruning} or knowledge distillation~\cite{gou2021knowledge}, as these methods also drop visual fidelity. In particular, we observe that naively shrinking STDiT~\cite{opensora} leads to significant visual loss, caused by iterative denoising steps, where errors propagate and accumulate to the final video. Another practical difficulty in achieving lightweight model optimization is the lack of resources required for model optimization; both re-training and fine-tuning state-of-the-art diffusion-based models typically demand several tens of GPUs, and the available datasets are often limited to effectively apply optimization methods. To tackle these challenges, we propose model training-free acceleration techniques for video generation in this work, \ie, Linear Proportional Leap (Sec. \ref{sec:ours1}) and Temporal Dimension Token Merging (Sec. \ref{sec:ours2}).

\parlabel{Straightness Constraints.} For video generation models that do not exhibit straightness during their denoising procedures, such as CogVideoX~\cite{yang2024cogvideox}, which employs DPM-Solver~\cite{zheng2023dpm}, Linear Proportional Leap (LPL) is currently inapplicable. In such cases, as described in previous work~\cite{ye2024schedule}, an alternative approach involves predicting modifications to the model's noise schedule by employing additional methodologies, such as reinforcement learning. However, these kind of methods require additional training  which may invoke extra costs, and necessitates retraining whenever the target model architecture changes. By contrast, LPL is a plug-and-play algorithm that can be applied in a model-agnostic manner, as long as a sufficient level of straightness is maintained during the denoising process, thereby eliminating the need for additional training. In future work, we plan to investigate the optimal reduction point achievable with this algorithm, and we anticipate that it will evolve into a widely applicable methodology for all Rectified Flow-based models~\cite{liu2022flow}.

\end{document}